\documentclass[a4paper]{cas-sc}
\ExplSyntaxOn
\exp_args:NNno \exp_args:Nno \use:n { \cs_gset:Npn \__make_fig_caption:nn #1#2 }
  {
    \exp_after:wN \use_ii_i:nn \exp_after:wN
      { \__make_fig_caption:nn {#1} {#2} }
      { \dim_set:Nn \l_fig_width_dim \linewidth }
  }
\exp_args:NNno \exp_args:Nno \use:n { \cs_gset:Npn \__make_tbl_caption:nn #1#2 }
  {
    \exp_after:wN \use_ii_i:nn \exp_after:wN
      { \__make_tbl_caption:nn {#1} {#2} }
      { \dim_set:Nn \l_tbl_width_dim \linewidth }
  }
\ExplSyntaxOff

\usepackage[numbers]{natbib}
\usepackage{subfigure}
\usepackage[justification=centering]{caption}
\usepackage{amsmath}
\usepackage{amsthm,amssymb}
\usepackage{arydshln}
\usepackage{verbatim}
\usepackage{graphicx}
\usepackage{subfigure}
\usepackage{booktabs}
\usepackage{amsfonts}
\usepackage{bm}
\usepackage{multirow}
\usepackage{enumerate}

\usepackage{float}
\usepackage{color}
\renewcommand*\ttdefault{txtt}
\usepackage{url}
\usepackage{setspace}
\usepackage{ulem}

\makeatletter
\renewcommand{\fnum@figure}{Fig. \thefigure.\@gobble}
\makeatother
                       
\def\tsc#1{\csdef{#1}{\textsc{\lowercase{#1}}\xspace}}
\tsc{WGM}
\tsc{QE}
\tsc{EP}
\tsc{PMS}
\tsc{BEC}
\tsc{DE}
\begin{document}
\begin{sloppypar}
\let\WriteBookmarks\relax
\def\floatpagepagefraction{1}
\def\textpagefraction{.001}
\shorttitle{MS-RNN: A Flexible Multi-Scale Framework for Spatiotemporal Predictive Learning}
\shortauthors{Ma et~al.}

\title [mode = title]{MS-RNN: A Flexible Multi-Scale Framework for Spatiotemporal Predictive Learning}   
%\tnotemark[1,2]

%\tnotetext[1]{This document is the results of the research
%   project funded by the National Science Foundation.}

%\tnotetext[2]{The second title footnote which is a longer text matter
%   to fill through the whole text width and overflow into
%   another line in the footnotes area of the first page.}

\author[1]{Zhifeng Ma}[orcid=0000-0002-0346-5757]
\ead{20b903027@stu.hit.edu.cn}
%\fnmark[1]
%\ead{cvr_1@tug.org.in}
%\ead[url]{www.cvr.cc, cvr@sayahna.org}

%\credit{Conceptualization of this study, Methodology, Software}

\address[1]{Faculty of Computing, Harbin Institute of Technology, Harbin, China}

\author[1]{Hao Zhang}[orcid=0000-0002-6769-2115]
\cormark[1]
\ead{zhh1000@hit.edu.cn}

\author[2]{Jie Liu} %[% role=Co-ordinator, suffix=Jr,]
%\fnmark[2]
%\ead{cvr3@sayahna.org}
%\ead[URL]{www.sayahna.org}

%\credit{Data curation, Writing - Original draft preparation}

\address[2]{International Research Institute for Artificial Intelligence, Harbin Institute of Technology, Shenzhen, China}
\ead{jieliu@hit.edu.cn}

\cortext[cor1]{Corresponding author}
%\cortext[cor2]{Principal corresponding author}
%\fntext[fn1]{This is the first author footnote. but is common to third
%  author as well.}
%\fntext[fn2]{Another author footnote, this is a very long footnote and
%  it should be a really long footnote. But this footnote is not yet
%  sufficiently long enough to make two lines of footnote text.}

%\nonumnote{This note has no numbers. In this work we demonstrate $a_b$
%  the formation Y\_1 of a new type of polariton on the interface
%  between a cuprous oxide slab and a polystyrene micro-sphere placed
%  on the slab.
%  }

\begin{abstract}
Spatiotemporal predictive learning, which predicts future frames through historical prior knowledge with the aid of deep learning, is widely used in many fields. Previous work essentially improves the model performance by widening or deepening the network, but it also brings surging memory overhead, which seriously hinders the development and application of this technology. In order to improve the performance without increasing memory consumption, we focus on scale, which is another dimension to improve model performance but with low memory requirement. The effectiveness has been widely demonstrated in many CNN-based tasks such as image classification and semantic segmentation, but it has not been fully explored in recent RNN models. In this paper, learning from the benefit of multi-scale, we propose a general framework named Multi-Scale RNN (MS-RNN) to boost recent RNN models for spatiotemporal predictive learning. We verify the MS-RNN framework by thorough theoretical analyses and exhaustive experiments, where the theory focuses on memory reduction and performance improvement while the experiments employ eight RNN models (ConvLSTM, TrajGRU, PredRNN, PredRNN++, MIM, MotionRNN, PredRNN-V2, and PrecipLSTM) and four datasets (Moving MNIST, TaxiBJ, KTH, and Germany). The results show the efficiency that RNN models incorporating our framework have much lower memory cost but better performance than before. Our code is released at \url{https://github.com/mazhf/MS-RNN}.
\end{abstract}

\iffalse
\begin{graphicalabstract}
\includegraphics{figs/grabs.pdf}
\end{graphicalabstract}

\begin{highlights}
\item Research highlights item 1
\item Research highlights item 2
\item Research highlights item 3
\end{highlights}
\fi


\maketitle

\section{Introduction}
Spatiotemporal predictive learning, also known as video prediction learning, has aroused widespread research interest in the field of deep learning and computer vision. The applications of this technique extend to many aspects of human life, including but not limited to precipitation nowcasting~\cite{shi2015convolutional, shi2017deep, wang2019memory, sonderby2020metnet, ma2022focal, luo2022predrann}, traffic and transportation~\cite{xu2018predcnn, zhang2017deep, zhang2022dynamic}, robot action prediction~\cite{finn2017deep, ebert2017self}, and pedestrian path prediction~\cite{wang2017predrnn, wang2018eidetic, wang2018predrnn++, lin2020self}. However, spatiotemporal prediction is an extremely challenging task, since motion always changes dramatically in both spatial and temporal domains~\cite{chang2022stam}. Therefore, how effectively learn complex spatiotemporal transformations is the key issue for improving the model's performance. 

\begin{figure}[t]
            \centering
    	\makebox[50mm]{
    	\includegraphics[width=0.72\linewidth]{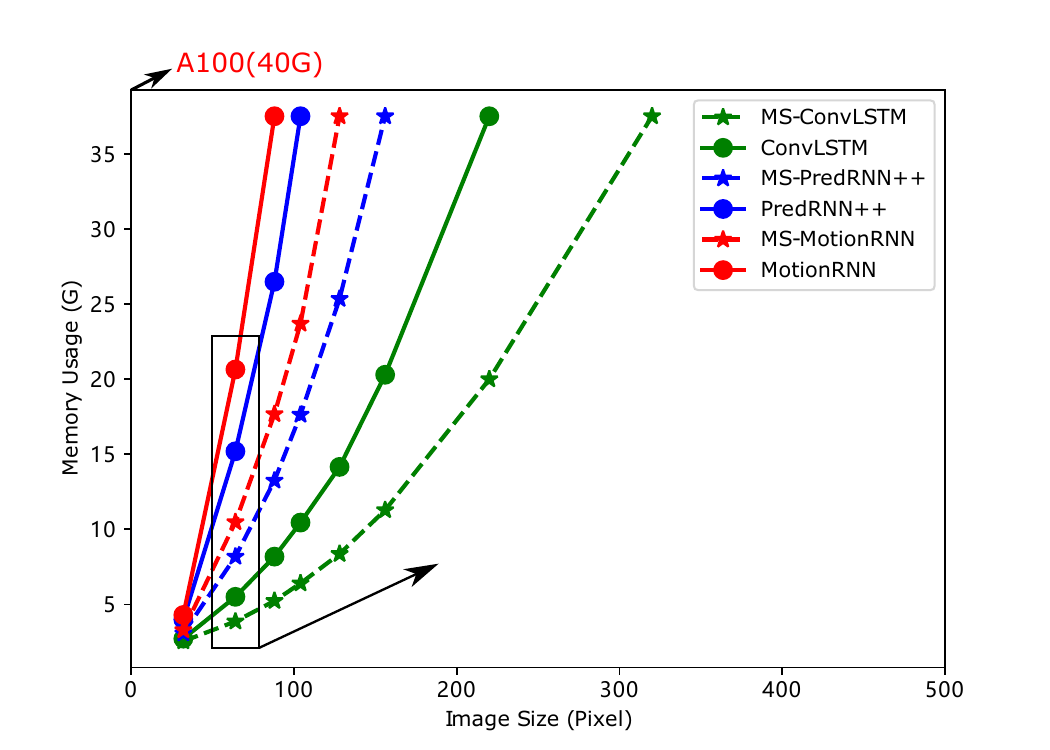}
    	\hspace{-70mm}
    	\resizebox{0.34\linewidth}{!}{
    		\begin{tabular}[b]{lcccc}
    			\toprule
    			%\hline
    			\centering
                Models & Memory$\downarrow$ & $\bigtriangleup$ & MSE$\downarrow$ & $\bigtriangleup$ \\
                \midrule
                ConvLSTM~\cite{shi2015convolutional} &  5.50G  & $-$ & 82.01 & $-$ \\
                \textbf{MS-ConvLSTM} & \textbf{3.86G} & \textbf{-29.8\%} & \textbf{58.31} & \textbf{-28.9\%} \\
                \midrule
                PredRNN++~\cite{wang2018predrnn++} & 15.20G & +176.4\% & 61.36 & -25.2\% \\
                \textbf{MS-PredRNN++} & \textbf{8.18G} & \textbf{+48.7\%} & \textbf{46.55} & \textbf{-43.2\%} \\
                \midrule
                MotionRNN~\cite{wu2021motionrnn} & 20.65G & +275.5\% & 55.35 & -32.5\% \\
                \textbf{MS-MotionRNN} & \textbf{10.48G} & \textbf{+90.5\%} & \textbf{51.55} & \textbf{-37.1\%} \\
                \bottomrule
                \multicolumn{3}{l}{\textbf{* Dataset: Moving MNIST}} \\
                \multicolumn{3}{l}{\textbf{* Image size: 64 $\times$ 64}\vspace{12.5mm}} \\
                \\
    		\end{tabular}
    	}
    }
	\caption{Comparison of memory usage and performance of RNN and MS-RNN. Given a fixed image size, the memory footprint of advanced models (e.g., ConvLSTM $\rightarrow$ PredRNN++ $\rightarrow$ MotionRNN) is getting larger and larger. On the contrary, our proposed multi-scale framework can greatly reduce their memory footprint and brings additional improvement. Meanwhile, for a fixed memory footprint, our framework can make the basic models handle larger images, which expands the serviceable scope of the basic models.}
	\label{fig:ms-rnn:cost}
	\vspace{-0.5cm}
\end{figure}

Since ConvLSTM~\cite{shi2015convolutional} successfully learns the changes of the target object in space and time simultaneously by bringing convolutional operations into LSTM~\cite{hochreiter1997long}, plenty of subsequent works have extended ConvLSTM by introducing well-designed modules. The networks of these followers become wider (e.g., TrajGRU~\cite{shi2017deep}, PredRNN~\cite{wang2017predrnn}, MIM~\cite{wang2019memory}, and PrecipLSTM~\cite{ma2022preciplstm}) and deeper (e.g., PredRNN++~\cite{wang2018predrnn++} and MotionRNN~\cite{wu2021motionrnn}). The result is a marginal improvement in model performance yet a significant increase in the usage of GPU memory. For example in Fig. \ref{fig:ms-rnn:cost}, compared with ConvLSTM, PredRNN++ reduces mean square error (MSE) by 25.2\% but increases the memory usage by 176.4\%. MotionRNN reduces MSE by 32.5\% but consumes much more memory by 275.5\%. However, the memory dedicated to training the neural network is limited and precious. 40 G memory is the limit of most modern GPUs~\cite{cheng2022stochastic}. The image size they can cope with is gradually decreasing when allowed to occupy full memory. As illustrated in Fig.~\ref{fig:ms-rnn:cost}, ConvLSTM can handle images with a maximum size of $220 \times 220$, while MotionRNN can only handle images with a maximum size of $80 \times 80$. On the contrary, real-world applications typically require high-resolution results~\cite{ravuri2021skilful, oprea2020review}. For example, the size of the radar image is usually over $500 \times 500$ in precipitation nowcasting, and the resolution of the front camera is larger than $1000 \times 1000$ pixels in the autonomous driving scenario. To fill this gap, the data has to be downsampled to reduce the memory footprint of the model in practice, which inevitably leads to the loss of details of the predicted frame and cannot meet the application requirements. Therefore, widening or deepening the model to improve the performance is not an optimal choice, since it leads to a sharp increase in memory consumption. 

We aim to improve the performance without increasing memory consumption. Depth and width are not the only dimensions that can enhance performance in computer vision. Researchers have recently identified two other dimensions, including cardinality~\cite{xie2017aggregated} and scale~\cite{gao2019res2net}. Cardinality groups convolutions to reduce the number of parameters before increasing the number of groups to boost performance. Although cardinality does not change the number of parameters, it increases GPU memory usage similar to depth and width. In order to obtain multi-scale representations of objects of different scales, visual feature extractors need to equip with a large range of receptive fields~\cite{gao2019res2net}. The way to expand the receptive field usually includes using a larger convolution kernel~\cite{szegedy2015going}, stacking deeper layers~\cite{he2016deep}, and inserting pooling (downsampling) layers~\cite{mathieu2016deep}, among which pooling is the most efficient, as it only adds layers without parameters while others either add width or depth. Furthermore, the most important thing is that pooling enables models to operate at lower resolutions, which leads to lower training overhead. Multi-scale techniques have been fully exploited in common vision tasks, such as ResNet~\cite{he2016deep} in classification, UNet~\cite{ronneberger2015u} in segmentation, 
Faster R-CNN~\cite{ren2015faster} in detection, MIMO-UNet~\cite{cho2021rethinking} in super-resolution, etc.  Regrettably, there are few models that effectively apply multi-scale structures to recurrent neural networks (RNNs) for predicting future frames. What leads to this situation is the complexity and difficulty of video prediction tasks. 

In this paper, we re-enable multiscale technology and propose a general framework named multi-scale RNN (MS-RNN) to improve existing RNN models. It should be emphasized that things are not as simple as directly migrating the multi-scale structures from convolutional models. First, we rearrange the tensor propagation flow of these increasingly complex RNN models and integrate them into a unified framework. Then we perform downsampling and upsampling processing for the tensors propagating along layers and times axes to get the multi-scale representations. Finally, we add skip connections between same-scale layers to get the MS-RNN framework. In addition, we also analyze the reasons for the overhead reduction and performance improvement brought by multi-scale structure: mainly due to the reduction of outputs memory and increase of the spatiotemporal receptive field. MS-RNN performs excellently in the experiments. For instance, as shown in Fig. \ref{fig:ms-rnn:cost}, compared with MotionRNN, MS-MotionRNN (MotionRNN using the MS-RNN framework) has better performance (reduce MSE from 55.35 to 51.55), while has much less memory overhead (reduce memory from 20.65G to 10.48G), which is almost half of vanilla MotionRNN. In addition, the performance of MS-ConvLSTM approaches the advanced MotionRNN. In practical applications, MS-ConvLSTM can be used to deal with high-resolution tasks (This paper adopts an RNN structure of 64 channels and 6 layers, which can be appropriately reduced to reduce memory usage to adapt to higher-resolution tasks (Section \ref{Analysis of Memory Reduction})). In summary, our contributions are as follows:

\begin{itemize}

\item We propose a general multi-scale framework named MS-RNN, which can simultaneously optimize the memory overhead and performance of RNNs.  %It is compatible with plenty of existing popular models such as ConvLSTM, TrajGRU, PredRNN, PredRNN++, MIM, MotionRNN, PredRNN-V2, and PrecipLSTM. 

\item We theoretically analyze reasons for the memory reduction and performance improvement brought by the multi-scale technique. 

\item We verify the effectiveness of MS-RNN on four datasets and eight RNNs. Experiments show that RNNs with our framework have much lower memory costs but higher performance than before.

\end{itemize}

\begin{table}[htbp]
    \centering
	\caption{Some representative video prediction works.}
	\resizebox{0.5\linewidth}{!}{
		\centering
		\begin{tabular}{lccccc}
			\toprule
			%\hline
			\centering
            Models & Publications & Basics & Characters \\
            \midrule
            ConvLSTM~\cite{shi2015convolutional} & NeurIPS 2015 & LSTM & Spatiotemporal  \\
            TrajGRU~\cite{shi2017deep} & NeurIPS 2017 & ConvLSTM & Wider  \\
            PredRNN~\cite{wang2017predrnn} & NeurIPS 2017 & ConvLSTM & Wider  \\
            PredRNN++~\cite{wang2018predrnn++} & ICML 2018 & ConvLSTM & Deeper  \\
            MIM~\cite{wang2019memory} & CVPR 2019 & ConvLSTM & Wider  \\
            MotionRNN~\cite{wu2021motionrnn} & CVPR 2021 & ConvLSTM & Deeper  \\
            PrecipLSTM~\cite{ma2022preciplstm} & TGRS 2022 & ConvLSTM & Wider  \\
            PredRNN-V2~\cite{wang2022predrnn} & TPAMI 2022 & ConvLSTM & Wider  \\
            MoDeRNN~\cite{chai2022modernn} & ICASSP 2022 & ConvLSTM & Multiscale  \\
            CMS-LSTM~\cite{chai2022cms} & ICME 2022 & ConvLSTM & Multiscale  \\
            SimVP~\cite{gao2022simvp} & CVPR 2022 & UNet & Multiscale  \\
            Earthformer~\cite{gao2022earthformer} & NeurIPS 2022 & Transformer & Multiscale  \\
            \bottomrule
		\end{tabular}
	} 
	\label{table:ms-rnn:related work}
    \vspace{-0.4cm}
\end{table}

\section{Related Work} \label{related work}
In view of the large number of video prediction models, we select some classic models to represent them (Table ~\ref{table:ms-rnn:related work}). The specific classification of video prediction models and the details of multi-scale video prediction models are as follows:

\textbf{Categories of Video Prediction Models.}
Common video prediction models include convolutional neural networks (CNNs), recurrent neural networks (RNNs), stochastic generative networks, and attention-based networks. These different network designs imply different inductive biases (prior knowledge). The UNet~\cite{ronneberger2015u} family rules CNNs. They mainly shine in the precipitation nowcasting domain. Representative models are RainNet~\cite{ayzel2020rainnet}, STConvS2S~\cite{castro2021stconvs2s}, Broad-UNet~\cite{fernandez2021broad}, and SimVP~\cite{gao2022simvp}. However, learning temporal trends with convolutions is difficult since they are designed for extracting spatial features, which may lead to difficult training and poor performance of CNNs~\cite{shi2017deep, ravuri2021skilful, wang2022predrnn}. The ConvLSTM~\cite{shi2015convolutional} family rules RNNs, which are thriving and constantly growing new members, like TrajGRU~\cite{shi2017deep}, MCNet~\cite{villegas2017decomposing}, PredRNN~\cite{wang2017predrnn},  E3D-LSTM~\cite{wang2018eidetic}, PredRNN++~\cite{wang2018predrnn++}, CubicLSTM~\cite{fan2019cubic}, MIM~\cite{wang2019memory}, SA-ConvLSTM~\cite{lin2020self}, PhyDNet~\cite{guen2020disentangling}, \cite{lee2021video}, MAU~\cite{chang2021mau}, MotionRNN~\cite{wu2021motionrnn}, STRPM~\cite{chang2022strpm}, PrecipLSTM~\cite{ma2022preciplstm}, PredRNN-V2~\cite{wang2022predrnn}, and etc. Generative adversarial nets (GANs)~\cite{mathieu2016deep, ravuri2021skilful} and variational autoencoders (VAEs)~\cite{denton2018stochastic, akan2021slamp, wu2021greedy} respectively use adversarial training and Bayesian variational inference to increase the randomness of prediction, and the images they generate are more realistic and clear, which alleviates the blur caused by the average loss ($L_1$ or $L_2$) to a certain extent. Since ViT~\cite{dosovitskiy2020image} pioneered the migration of Transformer~\cite{vaswani2017attention} from natural language processing to computer vision, the Transformer family has also begun to settle in video prediction domain, such as VideoGPT~\cite{yan2021videogpt}, TATS~\cite{ge2022long}, Earthformer~\cite{gao2022earthformer}, and MaskViT~\cite{gupta2022maskvit}. Distinct from other kinds of models, RNNs introduce the correct spatiotemporal inductive bias at birth, which neither introduces tricks to sustain training, nor is difficult to achieve equilibrium and stability, nor is lack of inductive bias leading to data greed. Accordingly, the primary focus of this paper is to investigate RNN models for spatiotemporal predictive learning. %In addition, there are some works from other novel perspectives, such as transfer learning~\cite{yao2020unsupervised, pavlyuk2020transfer, han2021advancing} and multimodal learning~\cite{espeholt2022deep, fu2022tell, han2022show}, to tackle the tough video prediction problem.

\textbf{Multiscale Models in Video Prediction Task.} 
In fact, multi-scale architectures are commonly used in the video prediction domain, such as RainNet~\cite{ayzel2020rainnet}, \cite{mathieu2016deep}, FitVid~\cite{babaeizadeh2021fitvid}, MCnet~\cite{villegas2017decomposing}, and SVG~\cite{denton2018stochastic}. However, these models either only use convolutions to construct multi-scale structures, or use the LSTMs as bottleneck layers between convolutional encoders and decoders, or use the first two structures in GANs or VAEs. These are different from the architecture proposed in this paper, MS-RNN improves models that have stacked RNN blocks. Stacking RNN layers is more powerful than stacking convolutional layers or a mix of convolutional and LSTM layers. It is the reason why subsequent ConvLSTM variants use the stacked RNN structure, and why these models make improvements between RNN layers. The most similar to our work is \cite{ebert2017self}, but using convolution and deconvolution for downsampling and upsampling brings additional parameters and memory usage. In addition, the difference between using multi-scale and single-scale structures is not discussed, and experiments are only performed on ConvLSTM while our structure is compatible with a series of subsequent RNN models. In recent times, SimVP~\cite{gao2022simvp} and Earthformer~\cite{gao2022earthformer} have adopted multi-scale structures similar to UNet. Unfortunately, their network design flaws make them inferior to MS-RNN (Table~\ref{table:ms-rnn:non-rnn vs}). In addition, MoDeRNN~\cite{chai2022modernn} and CMS-LSTM~\cite{chai2022cms} are also dedicated to attaining adjacent contextual and multi-scale features in spatiotemporal predictive learning. Both of them obtain contextual features by repeatedly interacting the previous output state with the current input. However, they choose different roads to design multi-scale structures, MoDeRNN employs multi-scale convolution kernels while CMS-LSTM employs multi-patch attention. Regrettably, both suffer from inefficiency. Besides, the former utilizes exceedingly large convolution kernels, resulting in significant computational costs, while the latter displays quadratic complexity, limiting its usage to low-resolution scenarios. Surprisingly, when incorporating them into our architecture like other RNN models, their predictive performance and training cost are significantly improved, which again demonstrates the generality and effectiveness of our architecture (Table~\ref{table:ms-rnn:taxibj-cost}).

\section{Preliminaries}\label{preliminaries}
\subsection{Problem Formulation} \label{sec:ms-rnn:problem formulation}
Suppose we use tensor $X_t \in \mathbb{R}^{c \times h \times w}$ to represent a frame of the video, where $t$, $c$, $h$, and $w$ denotes time, channels, height, and width, respectively. Then a piece of video can be expressed as \{$X_0, ..., X_{m-1}, X_{m}, ..., X_{m+n-1}$\}, where $m+n$ is total frames of the video. The video prediction problem is to use the first $m$ frames $X=\{X_0, ..., X_{m-1}$\} to predict the next $n$ frames $Y=\{X_{m}, ..., X_{m+n-1}$\}, and the neural network parametrized by $\theta$ needs to maximize the following likelihood:
\begin{gather}\label{eq:ms-rnn:problem}
    \begin{split}
        & \theta^* = \mathop{\arg\max}\limits_{\theta} \; P(Y|X;\theta). \\
	\end{split}
\end{gather} 

\begin{table}[htbp]
    \centering
	\caption{The complexity of the convolution and ConvLSTM.}
	\resizebox{0.42\linewidth}{!}{
		\centering
		\begin{tabular}{lccccc}
			\toprule
			%\hline
			\centering
            Models & Params & FLOPs & Memory \\
            \midrule
            Convolution & $c^2k^2$ & $2bc^2hwk^2$ & $bchw$ \\
            ConvLSTM & $\tilde{U}c^2k^2$ & $2\tilde{U}bc^2hwk^2$ & $Ubchw$ \\
            \bottomrule
		\end{tabular}
	} 
	\label{table:ms-rnn:convlstm-complexity}
    \vspace{-0.4cm}
\end{table}

\subsection{ConvLSTM} \label{sec:ms-rnn:convlstm} 
We first show the formula of ConvLSTM~\cite{shi2015convolutional}, which is beneficial to understand how ConvLSTM works and why it is suitable for spatiotemporal prediction tasks. The formula of the ConvLSTM unit is 
\begin{equation}
    \begin{aligned}
    & i_t = \sigma(W_{ix} \ast X_{t}^l + W_{ih} \ast H_{t-1}^l), \\
    & f_t = \sigma(W_{fx} \ast X_{t}^l + W_{fh} \ast H_{t-1}^l), \\
    & g_t = \tanh(W_{gx} \ast X_{t}^l + W_{gh} \ast H_{t-1}^l), \\
    & C_{t}^l = f_t \odot C_{t-1}^l + i_t \odot g_t, \\
    & o_t = \sigma(W_{ox} \ast X_{t}^l + W_{oh} \ast H_{t-1}^l), \\
    & H_{t}^l = o_t \odot \tanh(C_{t}^l).
    \label{eq:ms-rnn:convlstm}
    \end{aligned}
\end{equation} 
The convolution and its parameters are denoted by $\ast$ and $W_{**}$, respectively. Other components belong to LSTM. The input, hidden state, and cell state of layer $l$ at time $t$ are represented by $X_t^l$, $H_t^l$, and $C_t^l$, respectively. The input gate, forget gate, input modulation gate, and output gate at time $t$ are stood for $i_t$, $f_t$, $g_t$, and $o_t$  respectively. $\sigma$, $\tanh$, and $\odot$ denote the sigmoid activation function, tanh activation function, and Hadamard product, respectively.

At first glance, the formula of ConvLSTM is almost the same as that of LSTM. The only difference is that the fully connected operation ($1\times1$ convolution) in LSTM is replaced by convolution (like $3\times3$ convolution). We know that LSTM is suitable for processing temporal tasks (sequences), and convolution is suitable for processing spatial tasks (images). The seamless fusion of convolution and LSTM will allow ConvLSTM to handle spatial and temporal tasks simultaneously, which is a genius invention. In addition, stacking multiple ConvLSTM units will enhance the model's temporal memory ability and expand the spatial receptive field, which is not available in LSTM, because stacking $1\times1$ convolutions cannot expand the receptive field~\cite{shi2015convolutional}.

Below we analyze the complexity of ConvLSTM. The complexity of ConvLSTM is mainly concentrated in convolution, and its complexity is relatively simple, which can be extended to ConvLSTM. Let the kernel size, batch size, height, width, and channels be denoted by $k$, $b$, $h$, $w$, and $c$, respectively. Suppose we study a layer of convolution and ConvLSTM with the same input and output channels (size). Overall, as presented in Table~\ref{table:ms-rnn:convlstm-complexity}. The parameters (parameter complexity), FLOPs (computation complexity), and memory (space complexity) of ConvLSTM are $\tilde{U}$, $\tilde{U}$, and ${U}$ times of convolution, respectively, where $\tilde{U}=8$ while ${U}=16$. Specifically, ConvLSTM contains 8 convolutions, so the parameters and FLOPs of ConvLSTM are 8 times that of convolution, where we ignore the FLOPs of operations other than convolution because they are relatively small~\cite{pfeuffer2019separable}. Particularly, we use the size of the convolution's forward output or backward gradient to represent the memory it occupies. Assuming that the activation layer and Hadamard product layer have the same size of memory as the convolutional layer, then $U=16$ (``+'' has nothing to occupy~\cite{sohoni2019low}). Further, for more complex ConvLSTM variants that use more convolutional gates, both $U$ and $\tilde{U}$ will increase together.

\iffalse
the following is our proof:
According to the chain rule, the calculation formula of the parameter $W_{ix}$ of ConvLSTM is:
\begin{normalsize}
\begin{equation}
    \begin{aligned}
    \frac{\partial L}{\partial W_{ix}} \! \! = \! \! \frac{\partial L}{\partial H_t^l} \! \frac{\partial H_t^l}{\partial C_t^l} \!( \! \frac{\partial C_t^l}{\partial i_t} \! ( \! \frac{\partial i_t}{\partial W_{ix}} \! + \! \frac{\partial i_t}{\partial H_{t-1}^l} \! )  \! \! + \! \! \frac{\partial C_t^l}{\partial C_{t-1}^l}\!( \! \frac{\partial C_{t-1}^l}{\partial i_{t-1}} \! \frac{\partial i_{t-1}}{\partial W_{ix}} \! \! + \! \! \frac{\partial C_{t-1}^l}{\partial C_{t-2}^l} \! ( \! \cdots \!) \!) \!)
    \end{aligned}
\end{equation}
\end{normalsize}
\fi

\begin{figure*}[t]
	\centering
	\subfigure[RNN Framework]{\includegraphics[width=0.3812\linewidth]{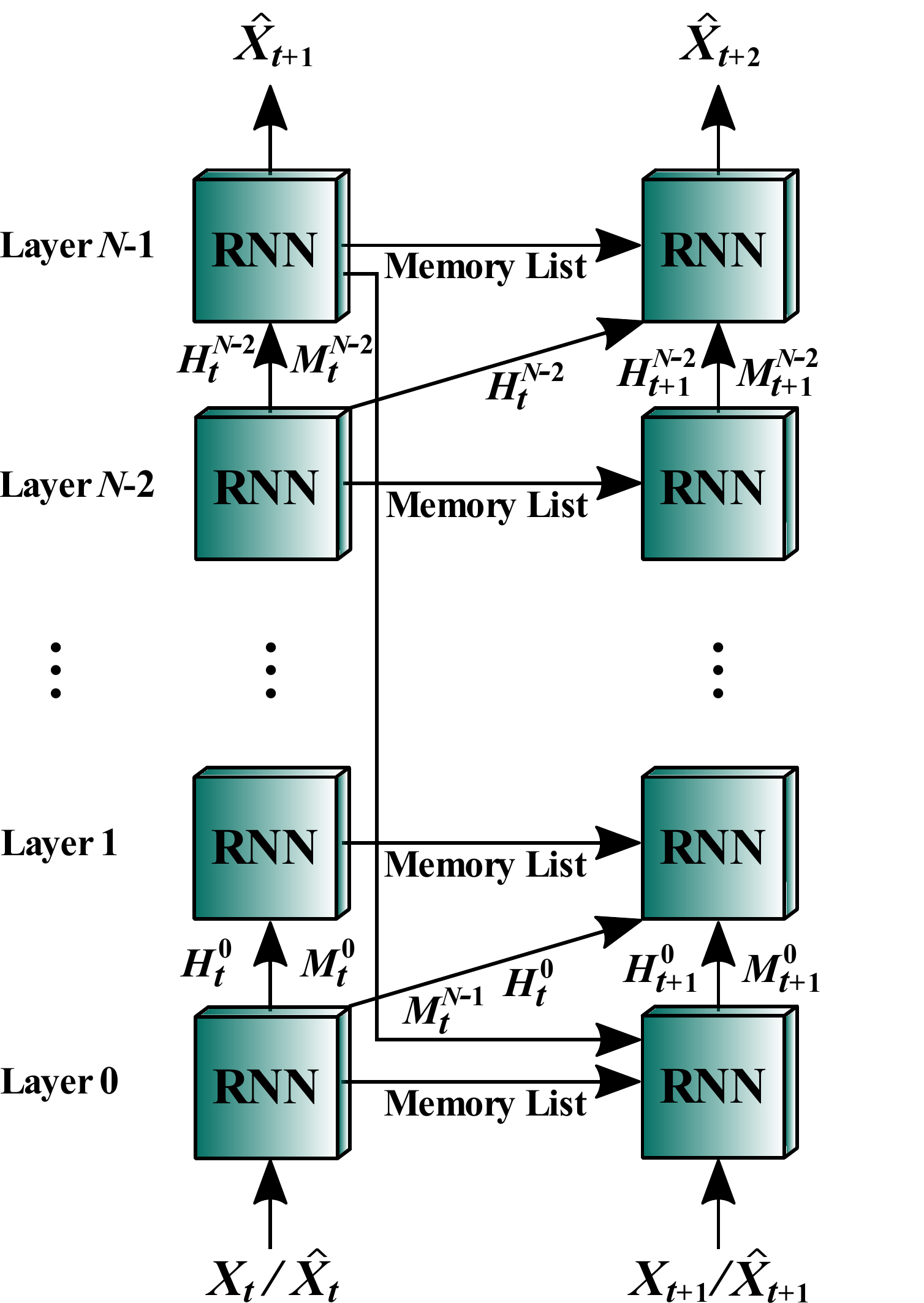} \label{fig:ms-rnn::rnn}}
	\subfigure[MS-RNN Framework]{\includegraphics[width=0.498\linewidth]{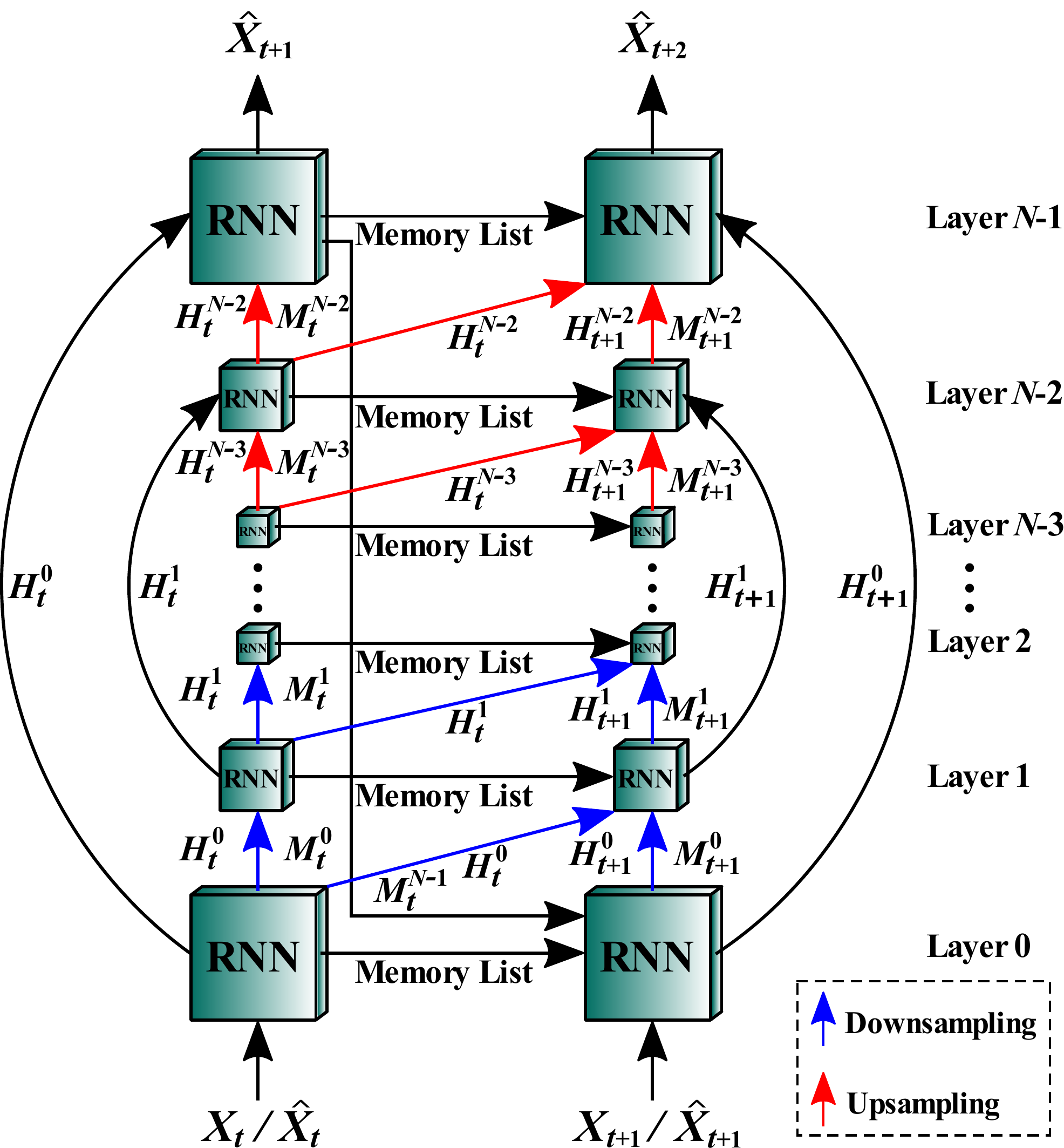} \label{fig:ms-rnn::ms-rnn}}
	\caption{First, we integrate ConvLSTM~\cite{shi2015convolutional}, TrajGRU~\cite{shi2017deep}, PredRNN~\cite{wang2017predrnn}, PredRNN++~\cite{wang2018predrnn++}, MIM~\cite{wang2019memory}, MotionRNN~\cite{wu2021motionrnn}, PredRNN-V2~\cite{wang2022predrnn}, and PrecipLSTM~\cite{ma2022preciplstm} into a unified RNN framework with the same layers (a), then we perform multi-scale processing to get the MS-RNN framework (b). Downsampling and upsampling operations are performed by maximum pooling (scale factor=2) and bilinear interpolation (scale factor=2) layers, respectively.}
	\label{fig:ms-rnn::rnn vs ms-rnn}
	\vspace{-0.5cm}
\end{figure*}

\section{MS-RNN}
\subsection{Framework} \label{sec:ms-rnn:framework}
Since the advent of ConvLSTM~\cite{shi2015convolutional}, many researchers have followed it, and they aim to solve the problem of the limited predictive ability of ConvLSTM. Unfortunately, although the successors are more powerful, the essential reason for the improvement is making ConvLSTM deeper or wider, which results in a sharp increase in memory usage. Meanwhile, scale is an efficient dimension to both improve the model performance and reduce memory usage~\cite{fan2021multiscale}, but it is rarely adopted by these ConvLSTM variants in recent years. Motivated by this, we construct a general multi-scale framework for these models by introducing multi-scale layers into them. To fit all the existing popular models, we make uniform modifications in two aspects. First, we unify their structure and tensor flow into one RNN framework (Fig.~\ref{fig:ms-rnn::rnn}). Then we embed the multi-scale structure into the unified framework, forming the MS-RNN architecture (Fig.~\ref{fig:ms-rnn::ms-rnn}). The structures of these models are similar but not totally the same. For example, PredRNN++ and MotionRNN have additional layers, and ConvLSTM and TrajGRU have no zigzag propagation path. We need to normalize them to make MS-RNN compatible with all models. Here we draw the most complex structure and the propagation path of all tensors in Fig.~\ref{fig:ms-rnn::rnn vs ms-rnn}. It only needs to remove part of the propagation path and tensors when applying to other relatively simple models.

\subsubsection{Struture Unification}
The unified RNN framework is shown in Fig.~\ref{fig:ms-rnn::rnn}, and different RNN models are compatible with this framework. Depending on the underlying model used, some tensors and propagation paths are optional. Specifically, in the vertical direction, the hidden state $H_t^l$ of time $t$ and layer $l$ propagates along the stacked ConvLSTM layers for predicting the frame at the next moment; in the horizontal direction, $H_t^l$ and the cell state $C_t^l$ convey the memory of ConvLSTM of the previous moment. TrajGRU~\cite{shi2017deep} combines optical flow and ConvGRU~\cite{shi2017deep} to dynamically model changes in pixels over time. Compared with ConvLSTM, it has only one memory $H_t^l$ propagating along the horizontal direction. Compared with ConvLSTM, PredRNN~\cite{wang2017predrnn} designs a new spatiotemporal memory cell $M_t^l$ that propagates along the vertical and horizontal direction with a zigzag trajectory. PredRNN++~\cite{wang2018predrnn++} rearranges the formulation of PredRNN and adds a gradient highway unit (GHU) between the first layer and the second layer, which we fuse with the second layer as one layer. Hence, the second layer of PredRNN++ has an additional memory $Z_t^l$ flowing along the horizontal direction. Compared with PredRNN, MIM~\cite{wang2019memory} decouples stationary and non-stationary information at the expense of adding a stationary memory cell ($S_t^l$) and a non-stationary memory cell ($N_t^l$), both of which propagate along the horizontal direction, and a diagonal propagation path is added for differential $H_{t}^{l-1}$ and $H_{t-1}^{l-1}$. MotionRNN~\cite{wu2021motionrnn} invents MotionGRU to decouple short-term and long-term motion and embed it between MIM layers. For MotionGRU, we fuse it with the upper MIM layer into one layer. Hence, Compared with MIM, MotionRNN adds two memories (transient memory $F_t^l$ and trend memory $D_t^l$) that propagate horizontally in each layer. PredRNN-V2~\cite{wang2022predrnn} extends PredRNN in two aspects: It proposes the reverse scheduled sampling strategy inspired by scheduled sampling~\cite{bengio2015scheduled}; It disentangles the  twisted memories cell state $C_t^l$ and spatiotemporal memory cell $M_t^l$. It's tensors and propagation paths are the same as PredRNN. Inspired by the first law of geography, PrecipLSTM~\cite{ma2022preciplstm} adds two memories $\mathrm{MS}_t^l$ and $\mathrm{MT}_t^l$ based on PredRNN to capture meteorological spatial and temporal changes respectively, both of which propagate along the horizontal direction.

In summary, all models have the tensor $H_t^l$ that propagates along the vertical direction. Except for ConvLSTM and TrajGRU, all models have the tensor $M_t^l$ that propagates along the zigzag direction. Only MIM and MotionRNN have the tensor $H_{t-1}^{l-1}$ that propagates along the diagonal direction. Spreading along the horizontal direction is the memory list. The memory list of TrajGRU is $[H_t^l]$. The memory list of ConvLSTM, PredRNN, and PredRNN-V2 is $[H_t^l, C_t^l]$. The memory list of the second layer of PredRNN++ is $[H_t^l, C_t^l, Z_t^l]$, and other layers are $[H_t^l, C_t^l]$. The memory list of MIM is $[H_t^l, C_t^l, N_t^l, S_t^l]$. The memory list of MotionRNN is $[H_t^l, C_t^l, N_t^l, S_t^l, F_t^l, D_t^l]$. The memory list of PrecipLSTM is $[H_t^l, C_t^l, \mathrm{MS}_t^l, \mathrm{MT}_t^l]$.

\subsubsection{Multi-scale Embedding}
Fig.~\ref{fig:ms-rnn::ms-rnn} depicts the MS-RNN framework. Unlike the RNN framework (Fig.~\ref{fig:ms-rnn::rnn}), downsampling (maximum pooling with scale factor of 2)) and upsampling (bilinear interpolation with scale factor of 2) layers are added to construct a mirror pyramid structure, and skip connections (use ``+'' to joint hidden states like $H_{t}^{0}$ and $H_{t}^{1}$) are added to combine the high-level semantic features and low-level detailed features of the same scales. It should be pointed out that all sampling layers of the same class are independent rather than shared operations. In the vertical direction, only downsampling and upsampling $H_t^l$ are required for MS-ConvLSTM and MS-TrajGRU. For MS-PredRNN, MS-PredRNN++, MS-MIM, MS-MotionRNN, MS-PredRNN-V2, and MS-PrecipLSTM, both $H_t^l$ and $M_t^l$ need to be sampled. In particular, for MS-MIM and MS-MotionRNN, sampling $H_{t-1}^{l-1}$ along the diagonal propagation is also required. During the flow of tensors along the vertical (layer) and horizontal (time) directions, memories in the memory list naturally maintain the same scale as the current layer, and no additional processing is required. For brevity, the skip connections in this paper only transmit the outputs of the basic RNN ($H_t^l$). In fact, the spatiotemporal memory $M_t^l$ (if any) can also be transmitted, which may bring additional performance improvements, but parameters and memory usage should be the same as MS-RNN. In addition, there are many choices of skip connection ways, such as similar to UNet 3+~\cite{huang2020unet} and fully connected ways, but experiments show that the one used in this paper is the simplest but the most effective (Table~\ref{table:ms-rnn:mnist-skip}). 

\subsection{Analysis of Memory Reduction} \label{Analysis of Memory Reduction}
There are three main types of memory consumption in training the neural network~\cite{sohoni2019low}, which are model memory, optimizer memory, and outputs memory. Model memory denotes the memory used to store the model parameters ($\mathcal{M}_{\mathrm{par}}$). Optimizer memory refers to the memory used to store the parameters gradients and any momentum buffers during training. For Adam~\cite{kingma2014adam} (the most popular optimizer used in spatiotemporal predictive learning), the optimizer memory is three times the size of the model memory (i.e., $3\mathcal{M}_{\mathrm{par}}$). Outputs memory denotes the memory occupied by the content of the forward and back propagations, which can be divided into forward outputs memory and backward outputs memory. Forward outputs memory ($\mathcal{M}_{\mathrm{out}}$) stores the outputs of each layer in the network for reuse in backpropagation. Backward outputs memory stores the gradients of outputs, whose size is equal to outputs~\cite{gao2020estimating}. Thus, the full memory overhead in training 
\begin{equation}
    \begin{aligned}
    \mathcal{M}_{\mathrm{all}} = 4\mathcal{M}_{\mathrm{par}} + 2\mathcal{M}_{\mathrm{out}}. 
    \end{aligned}
\end{equation}

Analogous to the analysis of the training cost of ConvLSTM in Section~\ref{sec:ms-rnn:convlstm}, we use $\tilde{U}$ and $U$ to denote multiples of RNN's parameter and space complexity relative to convolution respectively, use $(b, c, h, w)$ to denote the shape of the input or output tensor, and use $k$ to denote the convolution kernel size. Furthermore, we denote the number of stacked layers of RNN by $N$ (usually $N\geqslant3$ for acceptable performance), and the number of recursion of RNN by $R$ (total sequence length minus 1, $m+n-1$). Suppose the data type is the 32-bit floating-point number. For the scale-invariant RNN model (Fig.~\ref{fig:ms-rnn::rnn}), the model memory usage 
\begin{equation}
    \begin{aligned}
    \mathcal{M}_{\mathrm{par}}^0 = 32N\tilde{U}c^2k^2 \ {\mathrm{bit}}
    \end{aligned}
    \label{eq:ms-rnnmem_par}
\end{equation}
and the forward outputs memory usage
\begin{equation}
    \begin{aligned}
    \mathcal{M}_{\mathrm{out}}^0 = 32RNUbchw \ {\mathrm{bit}}.
    \end{aligned}
    \label{eq:ms-rnn out mem}
\end{equation}

For the scale-variant RNN model (Fig.~\ref{fig:ms-rnn::ms-rnn}), max pooling, bilinear interpolation, and skip connections (``+'') do not introduce extra parameters, and model parameters memory is independent of scale. Thus, the model parameters memory of MS-RNN $\mathcal{M}_{\mathrm{par}}^1$ is equal to $\mathcal{M}_{\mathrm{par}}^0$. Although skip connections (``+'') do not have output memory~\cite{sohoni2019low} while the pooling layer and the upsampling layer have forward and backward outputs memory~\cite{gao2020estimating}, MS-RNN only uses a few sampling operations, which can be ignored. If $N$ is odd, the forward outputs memory usage 
\begin{equation}
    \begin{aligned}
    \mathcal{M}_{\mathrm{out}}^1 & = 2\times32RUbc \left(hw + \frac{hw}{4} + \frac{hw}{16} + \cdots + \frac{hw}{4^{\frac{N-3}{2}}} \right)  \\
    & \hspace{4mm} + 32RUbc \times \frac{hw}{4^{\frac{N-1}{2}}} \\
    & =  \left(\frac{256}{3}\left(1 - {\left(\frac{1}{4}\right)}^{\frac{N-1}{2}}\right)  + \frac{32}{4^{\frac{N-1}{2}}}\right)RUbchw  \ {\mathrm{bit}},
    \end{aligned}
\end{equation}
and if $N$ is even, we have
\begin{equation}
    \begin{aligned}
   {M}_{\mathrm{out}}^1 & = 2\times32RUbc \left(hw + \frac{hw}{4} + \frac{hw}{16} + \cdots + \frac{hw}{4^{\frac{N-2}{2}}} \right) \\
    & = \frac{256}{3}\left(1 - {\left(\frac{1}{4}\right)}^{\frac{N}{2}}\right)RUbchw  \ {\mathrm{bit}}. \\
    \end{aligned}
\end{equation}
Both $\mathcal{M}_{\mathrm{\mathrm{out}}}^0$ and $\mathcal{M}_{\mathrm{out}}^1$ are monotonically increasing functions for $N$, and $\mathcal{M}_{\mathrm{out}}^1$ is always smaller than $\mathcal{M}_{\mathrm{out}}^0$. Actually, 
\begin{equation}
    \begin{aligned}
    & 96RUbchw \ {\mathrm{bit}} \leq  \mathcal{M}_{\mathrm{out}}^0 \leq +\infty \ {\mathrm{bit}}, \\
    & 72RUbchw \ {\mathrm{bit}} \leq \mathcal{M}_{\mathrm{out}}^1 \leq \frac{256}{3} \approx 85RUbchw \ {\mathrm{bit}}.
    \end{aligned}
\end{equation}
Compared to vanilla RNN, the memory overhead of MS-RNN is reduced by
\begin{equation}
    \begin{aligned}
    \frac{\mathcal{M}_{\mathrm{all}}^0  - \mathcal{M}_{\mathrm{all}}^1}{\mathcal{M}_{\mathrm{all}}^0} & = \frac{4\mathcal{M}_{\mathrm{par}}^0 + 2\mathcal{M}_{\mathrm{out}}^0 - 4\mathcal{M}_{\mathrm{par}}^1 - 2\mathcal{M}_{\mathrm{out}}^1}{4\mathcal{M}_{\mathrm{par}}^0 + 2\mathcal{M}_{\mathrm{out}}^0} \\
    &= \frac{\mathcal{M}_{\mathrm{out}}^0 - \mathcal{M}_{\mathrm{out}}^1}{2\mathcal{M}_{\mathrm{par}}^0 + \mathcal{M}_{\mathrm{out}}^0} \\
    &\overset{(*)}{\approx} \frac{\mathcal{M}_{\mathrm{out}}^0 - \mathcal{M}_{\mathrm{out}}^1}{\mathcal{M}_{\mathrm{out}}^0},
    \end{aligned}
\end{equation}
where ($*$) holds for the fact that the outputs memory ($2\mathcal{M}_{\mathrm{out}}$) is usually much larger than other memory ($4\mathcal{M}_{\mathrm{par}}$)~\cite{sohoni2019low}. 

The memory footprint reduction is significant. For $N \geq 3$, the theoretical reduction is at least 25\% ($N=3$). In practice, we usually use deeper networks for better performance. For example, suppose we use the RNN with 6 layers (i.e., $N=6$), the memory reduction 
\begin{equation}
    \begin{aligned}
    \frac{\mathcal{M}_{\mathrm{all}}^0  - \mathcal{M}_{\mathrm{all}}^1}{\mathcal{M}_{\mathrm{all}}^0} &= \frac{192RUbchw - 84RUbchw}{384\tilde{U}c^2k^2 + 192RUbchw} \\
    &= \frac{9RUbhw}{32\tilde{U}ck^2 + 16RUbhw} \\
    &\approx 56.25\%.
    \end{aligned}
\end{equation}
In reality, this ratio varies according to different experiment settings (Section~\ref{sec:ms-rnn:experiments}). Different basic RNN models and different datasets result in different $RUbhw$, which determines the magnitude of $\mathcal{M}_{\mathrm{out}}$ and whether $\mathcal{M}_{\mathrm{par}}$ can be ignored. When $RUbhw$ is larger, the ratio of memory reduction is closer to $56.25\%$. In other words, the longer the sequence, or the more complex the basic model, or the larger the batch size, or the larger the image size, the more effective MS-RNN is.

\iffalse
The memory consumption is quite different for neural network inference. The neural network has optimized parameters. Only a forward pass is necessary, and there is no backpropagation pass. In addition, the forward outputs memory can be reused~\cite{zhuang2021convolutional}. As a result, the full memory overhead in inference 
\begin{equation}
    \begin{aligned}
    \mathcal{M}_{\mathrm{all}}^* < \mathcal{M}_{\mathrm{par}} + \mathcal{M}_{\mathrm{out}} < \mathcal{M}_{\mathrm{all}}. 
    \end{aligned}
\end{equation}
This shows that as long as the model can be trained, it must be able to infer on the same device. It can even be deployed on edge devices. 

Model training is the first step, and model inference is the second step. If the model is too large, it can not be trained let alone inference, or can only be used for low-resolution inference. Therefore, memory usage during model training is critical. This is why we mainly analyze the memory of the training process. Since the situation of model inference is more complicated, we leave it to future work to analyze whether the multi-scale architecture can reduce memory usage and how much it can be reduced during inference.
\fi

\subsection{Analysis of FLOPs Reduction} \label{Analysis of FLOPs Reduction}
According to the analyses in Section~\ref{sec:ms-rnn:convlstm} and Section~\ref{Analysis of Memory Reduction}, we can easily get the training FLOPs of the scale-invariant RNN model 
\begin{equation}
    \begin{aligned}
    \mathcal{F}^0 &= 2RN\tilde{U}bc^2hwk^2,
    \end{aligned}
    \label{eq:ms-rnn rnn flops}
\end{equation}
and the training FLOPs (neglecting the FLOPs of few sampling and skip connection operations) of the scale-variant RNN model 
\begin{equation}
    \begin{aligned}
    \mathcal{F}^1 \! = \! 
    \begin{cases}
    \! \left( \! \frac{16}{3} \! \left(\! 1 \!-\! {\left(\frac{1}{4}\right)}^{\frac{N-1}{2}}\!\right) \! + \!\frac{2}{4^{\frac{N-1}{2}}}\!\right)\!\!R\tilde{U}bc^2hwk^2, \!\! & \text{if} \ N\!=\!\text{odd}, \\
    \! \frac{16}{3}\left(1 - {\left(\frac{1}{4}\right)}^{\frac{N}{2}}\right)R\tilde{U}bc^2hwk^2, \!\! & \text{if}\ N\!=\!\text{even}.
    \end{cases}
    \end{aligned}
\end{equation}
Both $\mathcal{F}^0$ and $\mathcal{F}^1$ are monotonically increasing functions for $N$, and $\mathcal{F}^1$ is always smaller than $\mathcal{F}^0$. Actually, 
\begin{equation}
    \begin{aligned}
    & 6R\tilde{U}bc^2hwk^2 \leq  \mathcal{F}^0 \leq +\infty , \\
    & \frac{9}{2}R\tilde{U}bc^2hwk^2 \leq \mathcal{F}^1 \leq \frac{16}{3} R\tilde{U}bc^2hwk^2 .
    \end{aligned}
\end{equation}
The FLOPs reduction is significant. For $N \geq 3$, the theoretical reduction is at least 25\% ($N=3$). In practice, we usually use deeper networks for better performance. For example, suppose we use the RNN with 6 layers (i.e., $N=6$), the FLOPs reduction 
\begin{equation}
    \begin{aligned}
    \frac{\mathcal{F}^0  - \mathcal{F}^1}{\mathcal{F}^0} &= \frac{12R\tilde{U}bc^2hwk^2 - \frac{21}{4}R\tilde{U}bc^2hwk^2}{12R\tilde{U}bc^2hwk^2} \\
    &= 56.25\%.
    \end{aligned}
\end{equation}

Obviously, the rate of FLOPs reduction is fixed and has nothing to do with many variables defined in this paper, which we have also verified in experiments (Table~\ref{table:ms-rnn:mnist-cost}).  
The variables that make up model memory $\mathcal{M}_{\mathrm{par}}$ and forward output memory $\mathcal{M}_{\mathrm{out}}$ also make up FLOPs, except for $U$, but $U$ and $\tilde{U}$ increase at the same time (Section~\ref{sec:ms-rnn:convlstm}). Thus, when the total training memory of the model increases, the training FLOPs must increases (Table~\ref{table:ms-rnn:mnist-cost}). Vice versa. Besides, compared with too large FLOPs, the consequences of too large memory may be more intuitive and more serious. Therefore, we focus on training memory analysis.

%Inference FLOPs?

\subsection{Analysis of Performance Improvement} 
As far as we know, there is little research conducted on the explainability of video prediction. \cite{huang2022understanding} uses ConvGRU~\cite{shi2017deep} as the encoder and CNN as the decoder to understand the spatiotemporal prediction model, where the encoder relies on a mechanism of extending the present and erasing the past to capture spatiotemporal dynamics, while the decoder synthesizes predictions from coarse to fine. Compared with RNN, there are more mature studies (\cite{luo2016understanding, ding2022scaling, gu2021interpreting}) on the interpretability of CNN. CNN can also be seen as a codec network, where convolution and pooling layers (optional) make up the encoder, while fully connected layers make up the decoder. They generally calculate and print the receptive field of the CNN's encoder. Given that the encoder of the convolutional RNN is similar to that of CNN, we also adopt this approach. As for the convolutional RNN decoder, our analysis has reached the same conclusion as \cite{huang2022understanding}, that is, decoding should be coarse to fine.

Limited by the scale of the convolution kernel, convolution can only capture the short-distance spatial dependence~\cite{mathieu2016deep}. To obtain richer contextual information, the simplest one is stacking more convolution RNN layers, another effective way is introducing pooling layers between convolutional RNN layers. Pooling makes the convolution kernel of the next layer see wider, i.e., a larger receptive field. However, pooling brings the loss of resolution, which has little impact on image or video classification tasks but is unacceptable for video prediction tasks, where the outputs and inputs have the same resolution. We solve this problem by using a mirror pyramid structure to restore resolution and join skip connections to preserve high-frequency information.

We analyze the effectiveness of MS-RNN by selecting ConvLSTM (Section~\ref{sec:ms-rnn:convlstm}) with six layers for instance. The initial three layers can function as an encoder to encode the current frame, while the final three layers can serve as a decoder to decode the following frame. Assume that both convolution and ConvLSTM use $3 \times 3$ convolution kernels. Assume that using the following functions will not change the size of the receptive field: element-wise addition ($+$), element-wise multiplication ($\odot$), and activation functions ($\sigma$ and $\tanh$). Then one layer of ConvLSTM and one layer of convolution (like $W_{*x} \ast X_{t}^l$) should have the same receptive field (Eq.~(\ref{eq:ms-rnn:convlstm})). Then the encoder of ConvLSTM will have a theoretical receptive field of $7 \times 7$. However, since sampling will result in a doubling of the receptive field~\cite{luo2016understanding}, the theoretical receptive field of the encoder of MS-ConvLSTM will reach $15 \times 15$. The above analysis shows that MS-ConvLSTM has a stronger ability in capturing faster spatiotemporal motion than ConvLSTM because it sees wider. Due to the adoption of a mirrored multi-scale architecture, the decoder of MS-ConvLSTM can gradually decode from small scale (coarse) to large scale (fine), which conforms to the natural law of from simple to difficult and step-by-step. In addition to gaining benefits from multi-scale decoding, MS-ConvLSTM's decoder also accepts high-frequency information of the same scale from its encoder, which avoids the loss of important information when stacking layers. These advantages are not available in the decoder of ConvLSTM. Overall, multi-scale codecs can capture the spatiotemporal motion of objects at different scales. Large scales can capture fine-grained changes, such as the movement of precipitation in local areas and the swing of arms and legs, while small scales can capture coarse-grained transformations, such as the shape of numbers and the outline of human beings.

In order to verify the above theoretical analysis, we perform corresponding experiments on the Moving MNIST~\cite{srivastava2015unsupervised} and KTH~\cite{schuldt2004recognizing} datasets (Section~\ref{sec:ms-rnn:recep and layer out} and Section~\ref{sec:ms-rnn:layer out}). Specifically, we visualize the gradients and receptive fields of the encoders of ConvLSTM and MS-ConvLSTM and print the output of each layer of the two RNNs. The final conclusion of theoretical analysis and experimental verification is that MS-RNN has a larger receptive field than RNN and synthesizes videos from coarse to fine, which accounts for the higher performance of MS-RNN than RNN.

\section{Experiments}\label{sec:ms-rnn:experiments}
We experiment on 4 video datasets, where Moving MNIST is synthetic while TaxiBJ~\cite{zhang2017deep}, KTH, and Germany~\cite{ayzel2020rainnet}) are realistic. Extensive experiments on these 4 datasets by 8 popular RNN models verify that our multi-scale architecture greatly reduces the memory footprint of the base models and improves their performance. On the one hand, the saved memory resource allows the models to handle larger images, which broadens the application of the models~\cite{wu2021greedy}. On the other hand, our framework enhances the performance of the base models, which will be of great help for practical applications such as human trajectory prediction and precipitation nowcasting. Additionally, we also verify that both multiple scales and skip connections ways like UNet play important roles in our framework.

\subsection{Implementation Details} \label{sec:ms-rnn:imple}
In this paper, identical experimental configurations are imposed on all RNN models, which can ensure the fairness of comparative experiments. All RNNs are composed of 6 layers of basic units. Adam~\cite{kingma2014adam} is used to optimize models, which employs an initial learning rate of 0.0003. We combine the $L_1$ and $L_2$ loss to optimize RNN models, which is better than just using a single one (Table~\ref{table:ms-rnn:loss vs}). During training, we run mini-batch gradient descent on 4 samples from all datasets at a time. The Germany dataset is trained for 16 epochs while the other datasets are trained for 20 epochs. The kernel size and channel of RNN are 3 (3, 5, and 7 for MoDeRNN~\cite{chai2022modernn}) and 64, respectively. We also enable scheduled sampling strategy~\cite{bengio2015scheduled} in the training process to guide the decoding future frames part of the model to gradually learn the sample sequence features and reduce the differences between the training and testing stages (reverse scheduled sampling for the encoding history frames part of PredRNN-V2~\cite{wang2022predrnn}). Our experimental equipment is a Linux server equipped with 2 CPUs and 4 GPUs, which has a total of 504G of physical memory. The system model of the server is Ubuntu 18.04 LTS, the CPU model is AMD EPYC 7532 with a base clock of 2.4GHz, and the GPU model is NVIDIA Tesla A100 with video memory of 40G. The experimental code for all RNNs involved in this article can be found in our code repository, which is released at \url{https://github.com/mazhf/MS-RNN}. 

\begin{table}[htbp]
    \begin{center}
	\caption{Training cost comparison on the Moving MNIST dataset. To speed up the training, we apply the distributed data parallel (DDP) technology of PyTorch~\cite{paszke2019pytorch} to a single machine with four graphics cards. Under the circumstance, each of them is responsible for computing gradients for a batch and they will automatically synchronize the gradients. The four graphics cards are load balanced with equal memory and FLOPs. The table below shows the training cost for one card, and it applies to other datasets as well. }
	\resizebox{0.55\linewidth}{!}{
		\centering
		% \begin{tabular}{l|c|cc|cc}
            \begin{tabular}{lccccc}
            \toprule
            Models & Params & Memory & $\bigtriangleup$ & FLOPs & $\bigtriangleup$ \\
            \midrule 
            ConvLSTM~\cite{shi2015convolutional}  & 1.77M & 5.50G & $-$ & 137.7G  & $-$ \\
            \textbf{MS-ConvLSTM} & 1.77M & \textbf{3.86G}  & \textbf{-29.8\%} & \textbf{60.3G} &	\textbf{56.21\%} \\
            \midrule
            PredRNN~\cite{wang2017predrnn} & 3.59M & 10.44G & $-$ & 279.3G & $-$ \\
            \textbf{MS-PredRNN} & 3.59M & \textbf{6.13G} & \textbf{-41.3\%}  &  \textbf{122.3G} &	\textbf{56.21\%} \\
            \midrule
            PredRNN++~\cite{wang2018predrnn++} & 5.61M & 15.20G & $-$ & 378.7G & $-$ \\
            \textbf{MS-PredRNN++} & 5.61M & \textbf{8.18G} & \textbf{-46.2\%} & \textbf{163.7G} &	\textbf{56.77\%} \\
            \midrule
            MIM~\cite{wang2019memory} & 7.36M & 19.69G & $-$ & 571.9G & $-$ \\
            \textbf{MS-MIM} & 7.36M & \textbf{10.09G} & \textbf{-48.8\%} & \textbf{250.4G} & \textbf{56.22\%} \\ 
            \midrule
            MotionRNN~\cite{wu2021motionrnn} & 7.56M &	20.65G & $-$ & 584.3G &	$-$ \\
            \textbf{MS-MotionRNN} & 7.56M &	\textbf{10.48G} &  \textbf{-49.2\%}  & \textbf{255.9G} &	\textbf{56.20\%} \\
            \midrule
            PredRNN-V2~\cite{wang2022predrnn} & 3.60M &	11.64G & $-$ & 283.1G &	$-$ \\
            \textbf{MS-PredRNN-V2} & 3.60M &	\textbf{6.66G} &  \textbf{-42.8\%}  & \textbf{124.0G} & \textbf{56.20\%} \\
            \midrule
            PrecipLSTM~\cite{ma2022preciplstm} & 7.04M &	26.36G & $-$ & 547.0G &	$-$ \\
            \textbf{MS-PrecipLSTM} & 7.04M & \textbf{12.56G} &  \textbf{-52.4\%}  & \textbf{239.5G} &	\textbf{56.22\%} \\
            \bottomrule
		\end{tabular}
	} 
	\label{table:ms-rnn:mnist-cost}
	\end{center}
    \vspace{-0.4cm}
\end{table}

\subsection{Moving MNIST}\label{sec:ms-rnn:mnist}
We follow the same procedure as \cite{srivastava2015unsupervised} to create Moving MNIST, which is consisted of 15,000 video clips with a training and test set ratio of 7:3. Each clip comprises 20 frames, with 10 frames dedicated to training and the remaining frames for prediction. The spatial resolution of each frame in the clip is 64 × 64, containing two randomly selected numbers from the static MNIST dataset~\cite{lecun1998gradient}. The numbers are displaced at random positions along with random speeds and directions before bouncing off the image border. Although the extrapolation method for each clip remains constant, it differs for different clips.

\begin{table}[htbp]
    \begin{center}
	\caption{Quantitative comparison on the Moving MNIST dataset.}
	\resizebox{0.4\linewidth}{!}{
		\centering
		\begin{tabular}{lccc}
            \hline
            Models & SSIM$\uparrow$ & MSE$\downarrow$ & MAE$\downarrow$ \\
            \hline
            ConvLSTM~\cite{shi2015convolutional} & 0.813  & 82.014 & 138.642\\
            \textbf{MS-ConvLSTM} & \textbf{0.863} &	\textbf{58.309} & \textbf{109.190}  \\
            \hline
            PredRNN~\cite{wang2017predrnn} & 0.851 & 64.782 & 114.326  \\
            \textbf{MS-PredRNN} & \textbf{0.884} & \textbf{48.312} & \textbf{94.431} \\
            \hline
            PredRNN++~\cite{wang2018predrnn++} & 0.856 	& 61.363 &	112.066 \\
            \textbf{MS-PredRNN++} & \textbf{0.892} &	\textbf{46.550} &	\textbf{89.307}\\
            \hline
            MIM~\cite{wang2019memory} & 0.861 &	58.634 &	109.885 \\
            \textbf{MS-MIM} & \textbf{0.884} 	& \textbf{48.593} &	\textbf{93.637}  \\ 
            \hline
            MotionRNN~\cite{wu2021motionrnn} & 0.869 &	55.350 &	104.096 \\
            \textbf{MS-MotionRNN} & \textbf{0.882} &	\textbf{51.550} &	\textbf{97.103}  \\
            \hline
            PredRNN-V2~\cite{wang2022predrnn} & 0.858 &	60.068 & 110.439 \\
            \textbf{MS-PredRNN-V2} & \textbf{0.884} &	\textbf{48.173} &	\textbf{94.973}  \\
            \hline
            PrecipLSTM~\cite{ma2022preciplstm} & 0.880 &	50.567 & 96.888 \\
            \textbf{MS-PrecipLSTM} & \textbf{0.885} &	\textbf{48.453} & \textbf{93.544}  \\
            \hline
		\end{tabular}
	} 
	\label{table:ms-rnn:mnist-dingliang}
	\end{center}
    \vspace{-0.3cm}
\end{table}

\subsubsection{Comparison of the Training Cost}  \label{training cost}
Table~\ref{table:ms-rnn:mnist-cost} depicts the training cost of different models. Consistent with our previous analysis, the multi-scale structure does not change the parameters of the basic RNN, but greatly reduces the memory and FLOPs occupancy of RNNs. The reduction of memory is variable, while the reduction of FLOPs is basically fixed. In detail, the more complex the model (larger $U$), the greater the memory reduction, i.e. the closer the reduction is to 56.25\%. In contrast, the reduction of FLOPs fluctuates around 56.25\%. Furthermore, due to the largely invariant nature of FLOPs, we only show results on one dataset. Later, the spotlight will be placed on the parameter and memory.

\begin{figure}[htbp]
	\centering
	\includegraphics[width=0.5\linewidth]{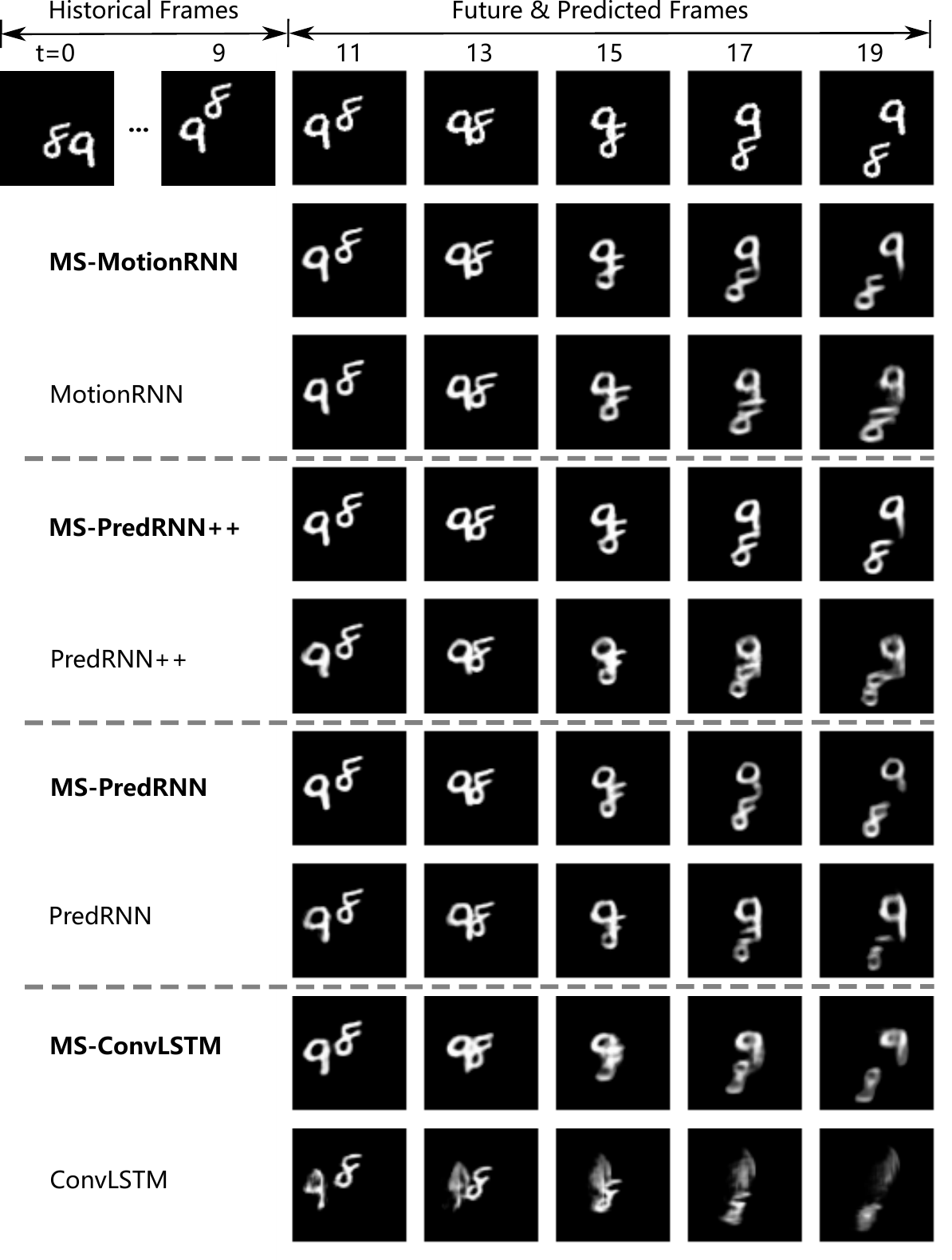}
	\caption{Qualitative comparison on the Moving MNIST dataset. The first row is the real frame, where the left is historical frames and the right is future frames. The other rows are predicted frames. This also applies to qualitative comparison on other datasets.}
	\label{fig:ms-rnn:mnist_demo}
	\vspace{-0.5cm}
\end{figure}

\subsubsection{Comparison of Performance}  \label{mnist-metrics}
The quantitative results are displayed in Table~\ref{table:ms-rnn:mnist-dingliang}. Comparative experiments on many basic RNN models have confirmed one point: our multi-scale structure can greatly improve the predictive ability of the basic RNN models. For example, compared to ConvLSTM, the MSE of MS-ConvLSTM is reduced by $28.9\%$ from 82.014 to 58.309. What's more surprising is that the mean performance of MS-ConvLSTM is even very close to that of the advanced model MotionRNN. In addition, among many multi-scale RNN models, MS-PredRNN++ has achieved the best results, and MS-PredRNN, MS-MIM, and MS-PredRNN-V2 compete with MS-PrecipLSTM for the second place. The models in the table are sorted by the year of publication, and we can find that the performance of the basic RNN model increases year by year, while the performance gap between the multi-scale RNN models is small. This indirectly shows that multi-scale skill is far more effective than deepening or widening the model and using complex structures or training strategies when applied to relatively simple tasks. It can make their performance quickly approach saturation.

The sample illustrated in Fig.~\ref{fig:ms-rnn:mnist_demo} presents a significant challenge due to the severe occlusion inside of the future sequence. The mean and variance of occlusion change rapidly, representing high-order non-stationarity~\cite{wang2019memory}. Still, MS-RNN produces clearer predictions, so we can conclude that multi-scale structure shows a strong ability in modeling non-stationary motion.

\begin{table}[htbp]
    \centering
	\caption{Comparison with competing multiscale non-RNN Models on the Moving MNIST dataset.}
	\resizebox{0.7\linewidth}{!}{
		\centering
		\begin{tabular}{lcccccc}
			\toprule
			%\hline
			\centering
            Models & Basics & Params & Memory & FLOPs & SSIM$\uparrow$ & MSE$\downarrow$ \\
            \midrule
            SimVP~\cite{gao2022simvp} & UNet & 43.2M & 4.4G & 68.5G & 0.734 & 129.2   \\
            Earthformer~\cite{gao2022earthformer} & Transformer & 6.7M & 38.8G & 519.5G & 0.787 & 71.0  \\
            MS-ConvLSTM & ConvLSTM & \textbf{1.8M} & \textbf{3.9G} & \textbf{60.3G} & 0.863 & 58.3 \\
            %MS-PrecipLSTM & RNN & 7.0M & 12.6G & 239.5G & 0.885 & 48.5 \\
            MS-PredRNN++ & ConvLSTM & 5.6M & 8.2G & 163.7G & \textbf{0.892} & \textbf{46.6} \\
            \bottomrule
		\end{tabular}
	} 
	\label{table:ms-rnn:non-rnn vs}
\end{table}

\subsubsection{Comparison with Competing Multiscale Non-RNN Models}  \label{non-rnn}
In 2022, some none RNN-based models are released, such as CNN-based model SimVP~\cite{gao2022simvp} and Attention-based model Earthformer~\cite{gao2022earthformer}. They all use a multi-scale architecture like UNet, where SimVP uses convolution to simultaneously capture space-time motion while Earthformer uses cuboid attention to model local and global space-time dynamics. Since SimVP and Earthformer use completely different types of neural networks than various RNNs studied in this paper, we do not adopt experimental settings in Section~\ref{sec:ms-rnn:imple} for them but adopt officially experimental settings obtained from their original papers and source code. Table~\ref{table:ms-rnn:non-rnn vs} compares them to our worst-performing model MS-ConvLSTM and best-performing model MS-PredRNN++. First, our models have fewer parameters than theirs. Second, the memory and FLOPs of MS-ConvLSTM are smaller than SimVP, while the memory and FLOPs of MS-PredRNN++ are smaller than Earthformer. Finally, both performed poorly in SSIM (structural similarity,~\cite{wang2004image})) and MSE metrics. In addition, SimVP suffers overfitting severely during the training process (2000 epochs), and Earthformer's overly complex model structure requires thousands of lines of code to reproduce, all of which are headaches.

\begin{figure*}[htbp]
	\centering
        \subfigure[The receptive field of the ConvLSTM encoder]{\includegraphics[width=0.49\linewidth]{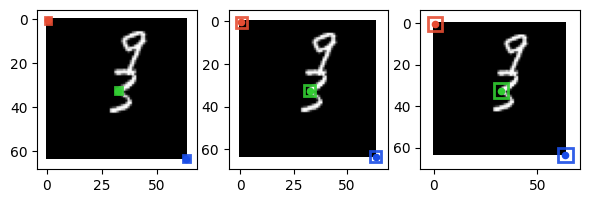} \label{fig:ms-rnn::rnn-recept}}
        \subfigure[The gradient of the ConvLSTM encoder]{\includegraphics[width=0.49\linewidth]{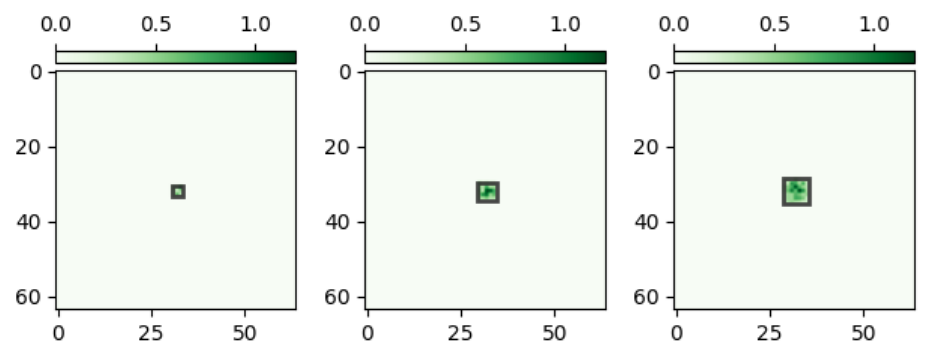}\label{fig:ms-rnn::rnn_grad}}
        \subfigure[The receptive field of the MS-ConvLSTM encoder]{\includegraphics[width=0.49\linewidth]{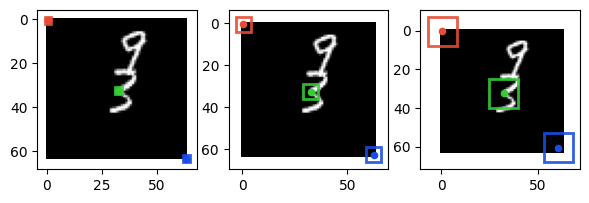} \label{fig:ms-rnn::ms-rnn-recept}}
        \subfigure[The gradient of the MS-ConvLSTM encoder]{\includegraphics[width=0.49\linewidth]{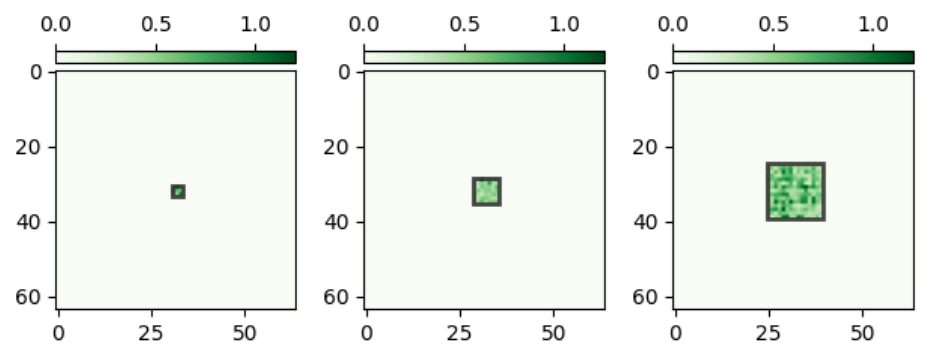}
        \label{fig:ms-rnn::ms-rnn_grad}} 
	\caption{The receptive field and gradient of the ConvLSTM encoder and MS-ConvLSTM encoder on the Moving MNIST dataset. In each subfigure from left to right are the first layer, the second layer, and the third layer of the encoder.}
	\label{fig:ms-rnn:recep and grad}
	\vspace{-0.5cm}
\end{figure*}

\begin{figure*}[htbp]
	\centering
	\includegraphics[width=1\linewidth]{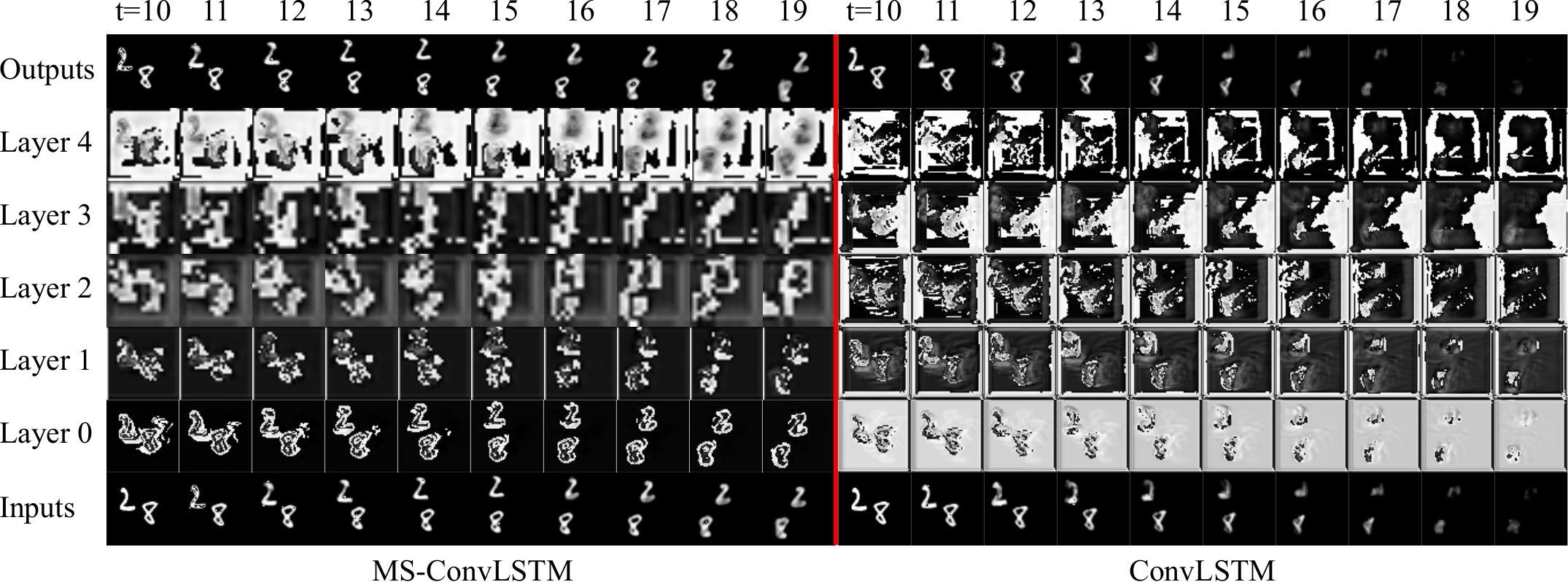}
	\caption{The layer outputs of ConvLSTM and MS-ConvLSTM on the Moving MNIST dataset.}
	\label{fig:ms-rnn:minist_layer_out}
	\vspace{-0.5cm}
\end{figure*}

\subsubsection{Visualization of Receptive Field and Layer Outputs}  \label{sec:ms-rnn:recep and layer out}
The receptive field refers to the size of the area where the point on the feature map of the neural network is mapped back to the input image. Simply put, it is the size of the input image that the feature map can see. By calculating the gradient of a point on the feature map relative to each point of the input image, the effective receptive field can be obtained. Then, the point with a gradient of 0 is in the blind field, otherwise, the point is in the receptive field. We employ an open-source toolkit (\url{https://github.com/shelfwise/receptivefield}) to visual receptive fields.

Fig.~\ref{fig:ms-rnn:recep and grad} shows the receptive field and gradient of the ConvLSTM encoder and MS-ConvLSTM encoder on the Moving MNIST dataset. With the stacking of ConvLSTM layers, the receptive fields of both ConvLSTM and MS-ConvLSTM are getting larger. However, the receptive field of the last layer of the MS-ConvLSTM encoder is significantly larger than that of ConvLSTM. Specifically, the receptive fields of the three layers of the ConvLSTM encoder are 3, 5, and 7, respectively, while the receptive fields of the three layers of the MS-ConvLSTM encoder are 3, 7, and 15, respectively. This phenomenon is due to the multiplication of the receptive field caused by the introduction of pooling layers~\cite{luo2016understanding}. Fig.~\ref{fig:ms-rnn::rnn-recept} and Fig.~\ref{fig:ms-rnn::ms-rnn-recept} show the receptive field of three positions in the feature map, which are the upper left corner, the center, and the lower right corner. Fig.~\ref{fig:ms-rnn::rnn_grad} and Fig.~\ref{fig:ms-rnn::ms-rnn_grad} show gradients of the center position in the feature map. From the comparison of the last receptive field maps of the two RNNs, we can see that MS-ConvLSTM almost sees the entire outline of the number ``3" (green rectangle), while ConvLSTM only sees part of the details of the number ``3" (green rectangle). In this way, MS-ConvLSTM will see more spatiotemporal context information, which makes capturing motion an easy task.

This conclusion is further supported by Fig.~\ref{fig:ms-rnn:minist_layer_out}, which displays the inputs, intermediate outputs of layers, and final outputs of ConvLSTM and MS-ConvLSTM. (i) Analysis for encoders (from inputs to layer 2): Firstly, the numbers in the layer features of both ConvLSTM and MS-ConvLSTM are gradually getting fatter. This indicates that their receptive fields increase as layers are stacked. Obviously, the numbers of MS-ConvLSTM are fatter than ConvLSTM in the same layers, that is, wider receptive fields. Second, we can identify numbers in the early features ($t=10, \dots, 15$) of ConvLSTM, while we cannot see numbers in the later features ($t=16, \dots, 19$) because of poor predictions (same for the decoder). Whereas, the features of the MS-ConvLSTM encoder gradually become abstract, and we cannot even tell that the numbers in layer 2 are 2 and 8. (ii) Analysis for decoders (from layer 3 to outputs): The decoder of MS-ConvLSTM makes predictions step by step, which is a coarse-to-fine process, where layer 3 predicts contours, layer 4 supplements details, and layer 5 (outputs) restores resolution. In contrast, we can always distinguish the numbers 2 and 8 from the early features of ConvLSTM's decoder, lacking this reasonable procedure. (iii) Analysis for models: In summary, MS-ConvLSTM rapidly expands the receptive field to capture spatiotemporal motion and gradually synthesizes future frames from coarse to fine. ConvLSTM slowly expands the receptive field and lacks a step-by-step frame synthesis procedure. Ultimately, MS-ConvLSTM outperforms ConvLSTM.

\begin{table}[htbp]
    \centering
	\caption{Training cost comparison and framewise comparison of MSE on the TaxiBJ dataset. The smaller MSE the better.}
	\resizebox{0.7\linewidth}{!}{
		\centering
		\begin{tabular}{lccccccc}
			\toprule
			%\hline
			\centering
            Models & Params & Memory & $\bigtriangleup$ & Frame 1 & Frame 2 & Frame 3 & Frame 4\\
            \midrule
            %\hline
            ConvLSTM~\cite{shi2015convolutional}  & 1.77M &	2.71G & $-$ & 0.187 &	0.252 &	0.300 &	0.334\\
            \textbf{MS-ConvLSTM} & 1.77M & \textbf{2.54G} & \textbf{-6.3\%} & \textbf{0.153} &	\textbf{0.195} & \textbf{0.236} & \textbf{0.278} \\
            \midrule
            TrajGRU~\cite{shi2017deep}  & 2.11M & 3.51G & $-$ & 0.179 &	0.255 &	0.306 &	0.350 \\
            \textbf{MS-TrajGRU} & 2.11M & \textbf{2.92G} & \textbf{-16.8\%} & \textbf{0.159} &	\textbf{0.212} &	\textbf{0.246} &	\textbf{0.282} \\
            \midrule
            PredRNN~\cite{wang2017predrnn} & 3.59M & 3.24G & $-$ & 0.160 &	0.201 &	0.241 &	0.274  \\
            \textbf{MS-PredRNN} & 3.59M & \textbf{2.82G} & \textbf{-13.0\%} & \textbf{0.145} &	\textbf{0.190} &	0.241 &	\textbf{0.273}\\
            \midrule
            PredRNN++~\cite{wang2018predrnn++} & 5.61M & 3.99G & $-$ & 0.157 	& 0.209 &	0.264 &	0.298 \\
            \textbf{MS-PredRNN++} & 5.61M &	\textbf{3.07G} & \textbf{-23.1\%} & \textbf{0.155} &	\textbf{0.208} &	\textbf{0.254} &	\textbf{0.294} \\
            \midrule
            MIM~\cite{wang2019memory} & 7.36M & 4.16G & $-$ & 0.152 &	0.202 &	0.250 &	0.287 \\
            \textbf{MS-MIM} &  7.36M &	\textbf{3.24G}  & \textbf{-22.1\%} & \textbf{0.151} &	\textbf{0.197} &	\textbf{0.244} &	\textbf{0.280}\\ 
            \midrule
            MotionRNN~\cite{wu2021motionrnn} & 7.56M & 4.29G & $-$ & 0.145 &	0.190 &	0.239 &	0.279  \\
            \textbf{MS-MotionRNN} & 7.56M & \textbf{3.28G} & \textbf{-23.5\%} & \textbf{0.142} & \textbf{0.185} & \textbf{0.219} & \textbf{0.251} \\
            \midrule
             CMS-LSTM~\cite{chai2022cms} & 4.48M & 6.87G & $-$ & 0.236 & 0.313 & 0.377 & 0.445\\
            \textbf{MS-CMS-LSTM} &  4.48M &	\textbf{4.33G}  & \textbf{-37.0\%} & \textbf{0.161} & \textbf{0.230} &	\textbf{0.286} &	\textbf{0.333}\\ 
            \midrule
            MoDeRNN~\cite{chai2022modernn} & 5.86M & 7.75G & $-$ & 0.161 & 0.214 & 0.262 & 0.298\\
            \textbf{MS-MoDeRNN} & 5.86M & \textbf{4.78G} & \textbf{-38.3\%} & \textbf{0.158} & \textbf{0.207} & \textbf{0.256} & \textbf{0.283}\\
            \bottomrule
		\end{tabular}
	} 
	\label{table:ms-rnn:taxibj-cost}
\end{table}

\begin{figure}[htbp]
	\centering
	\includegraphics[width=0.45\linewidth]{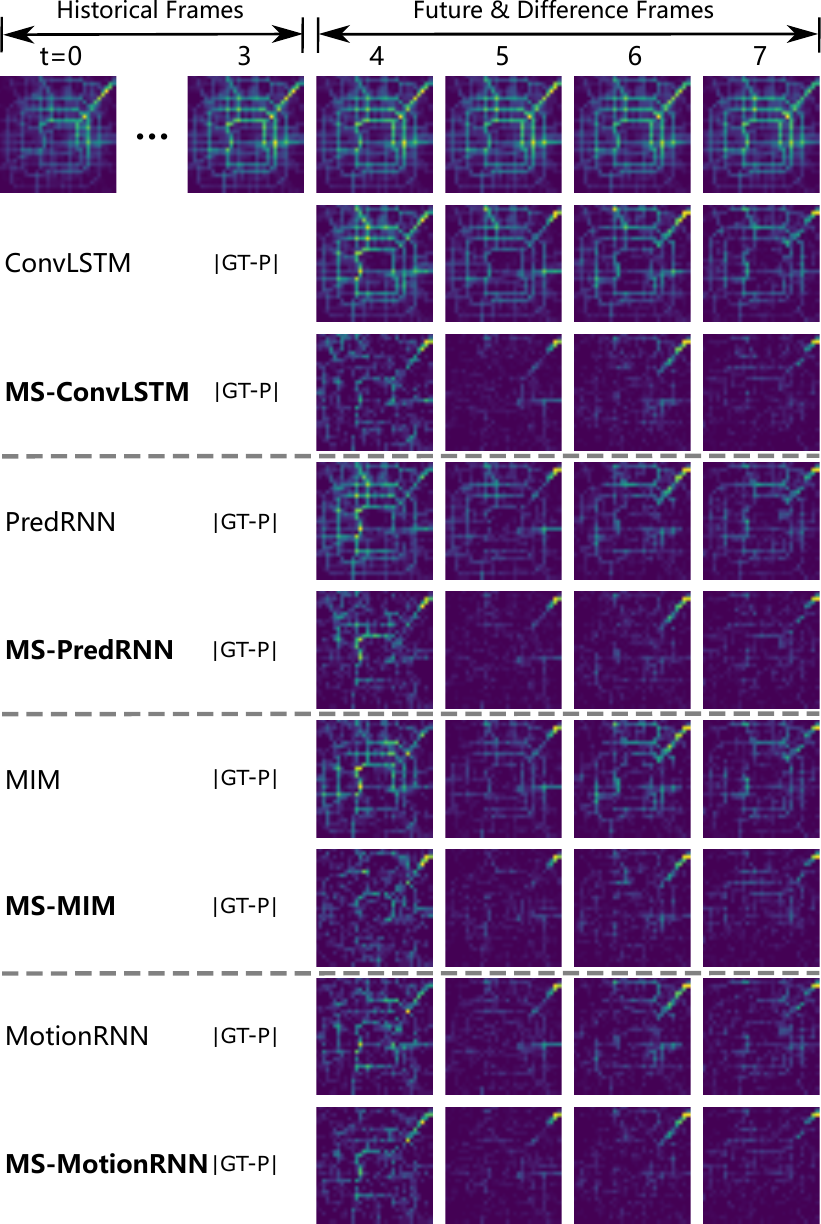}
	\caption{Qualitative comparison on the TaxiBJ dataset. $|$GT-P$|$ denotes the absolution difference frame between ground truth and predicted frames, prediction is the most accurate when there is nothing in the difference frame.}
	\label{fig:ms-rnn:taxibj_demo}
	\vspace{-0.1cm}
\end{figure}

\subsection{TaxiBJ Traffic Flow}\label{sec:ms-rnn:taxi}
The TaxiBJ dataset~\cite{zhang2017deep} comprises four-time periods of taxicab trajectory collected through sensors installed in various cars in Beijing. The size of each frame in the dataset is $32\times32$. We use 4 historical frames to predict 4 future frames, which is the traffic flow in the next two hours. Following the experimental settings employed in MIM~\cite{wang2019memory}, the dataset is divided into a training set with 19,560 samples and a test set with 1,344 samples.

\subsubsection{Comparison of the Training Cost and Performance}  \label{taxibj dingliang dingxing}
As shown in Table~\ref{table:ms-rnn:taxibj-cost}, the models that adopted multi-scale structures require less video card memory usage in the training phase but have higher accuracy than basic models in the test phase. The memory reduction of various RNN models is lower on TaxiBJ than on Moving MNIST due to the smaller image size ($h$ and $w$) of TaxiBJ and shorter overall sequence length ($R+1$). In addition, we also compare the absolute difference between the true images and the predicted images (Fig.~\ref{fig:ms-rnn:taxibj_demo}). Obviously, the introduction of multi-scale technology improves the predictive ability of the basic models and reduces the prediction error.

\subsubsection{Comparison with Competing Multiscale RNN Models}  \label{competing ms}
Multiscale RNN models that compete with MS-RNN are MoDeRNN~\cite{chai2022modernn} and CMS-LSTM~\cite{chai2022cms}, both published in 2022. They choose different roads to design multi-scale structures, MoDeRNN uses multi-scale convolution kernels while CMS-LSTM adopts multi-patch attention. First, it can be seen from Table~\ref{table:ms-rnn:taxibj-cost} that both competitors perform poorly, they do not even surpass MS-TrajGRU, which is our worst-performing model let alone the best-performing MS-MotionRNN. Second, although they have fewer parameters than MS-MotionRNN, their memory costs are more than twice that of MS-MotionRNN. This is caused by the square complexity of the attention of CMS-LSTM and the excessive convolution kernel (3, 5, and 7) of MoDeRNN. These multi-scale designs also cause them to only work in low-resolution scenarios. We run out of memory when launching them on other higher-resolution datasets. In conclusion, the multi-scale design of MoDeRNN and CMS-LSTM is expensive and inefficient. Dramatically, when we incorporate them into our multiscale architecture (improve them like ConvLSTM), both their memory and performance are improved significantly.

\begin{figure*}[t]
	\centering
	\includegraphics[width=1\linewidth]{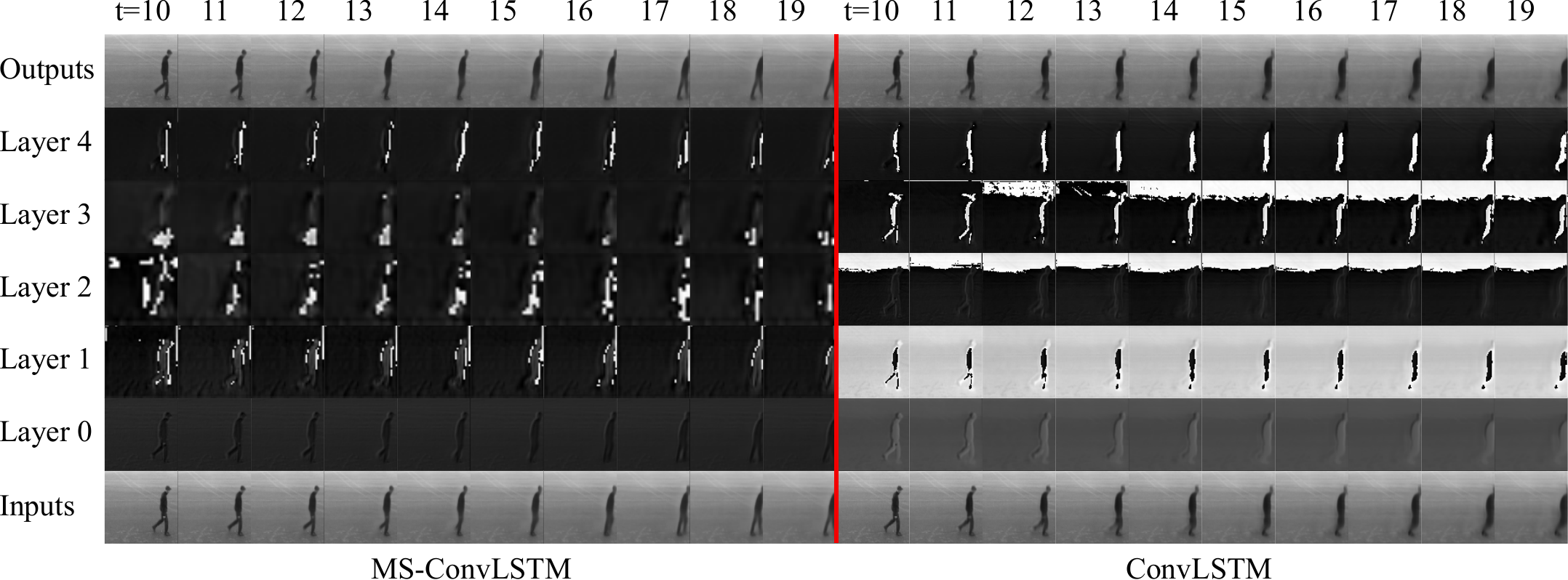}
	\caption{The layer outputs of ConvLSTM and MS-ConvLSTM on the KTH dataset. MS-ConvLSTM rapidly expands the receptive field to capture motion and gradually synthesizes future
frames from coarse to fine. ConvLSTM slowly expands the receptive field and lacks the natural synthesis
procedure.}
	\label{fig:ms-rnn:kth_layer_out}
	\vspace{-0.3cm}
\end{figure*}

\subsection{KTH Human Action}\label{sec:ms-rnn:kth}
The KTH dataset~\cite{schuldt2004recognizing} consists of six human actions, including boxing, hand clapping, hand waving, running, jogging, and walking, executed by 25 participants in four distinct scenarios, namely indoors, outdoors, outdoors with various clothing, outdoors with scale variations. For this study, we solely select actions that involve a significant range of motion, such as running, jogging, and walking. The video frames are resized to $88 \times 88$ pixels and the frames are cropped to ensure the human subjects are consistently visible. Individuals 1 to 16 are used for training, while those numbered 17 to 25 are used for testing. The sliding window for all actions is set to 20, where the first 10 frames are used to predict the next 10 frames. The stride for jogging and walking is set to be 10, while for running, it is 3.

\begin{table}[htbp]
    \centering
	\caption{Training cost comparison and quantitative comparison on the KTH dataset.}
	\resizebox{0.7\linewidth}{!}{
		\centering
		\begin{tabular}{lccccccc}
			\toprule
			%\hline
			\centering
            Models & Params & Memory & $\bigtriangleup$ & SSIM$\uparrow$ & PSNR$\uparrow$ & GDL$\downarrow$ & MSE$\downarrow$ \\
            \midrule
            %\hline
            ConvLSTM~\cite{shi2015convolutional}  & 1.77M &	8.35G & $-$ & 0.896 & 30.09 & 152.28 &	13.56\\
            \textbf{MS-ConvLSTM} & 1.77M & \textbf{5.17G} & \textbf{-38.1\%} & \textbf{0.917} &	\textbf{31.37} & \textbf{140.99} & \textbf{9.61} \\
            \midrule
            PredRNN~\cite{wang2017predrnn} & 3.59M & 18.43G & $-$ & 0.917 &	31.41 &	139.07 & 10.29 \\
            \textbf{MS-PredRNN} & 3.59M & \textbf{9.64G} & \textbf{-47.7\%} & \textbf{0.925} &	\textbf{32.34} &	\textbf{136.87} &	\textbf{8.08}\\
            \midrule
            PredRNN-V2~\cite{wang2022predrnn} & 3.60M & 20.26G & $-$ & 0.915 & 31.51 &	141.28 & 10.36 \\
            \textbf{MS-PredRNN-V2} & 3.60M &	\textbf{10.37G} & \textbf{-48.8\%} & \textbf{0.924} &	\textbf{32.19} &	\textbf{137.12} &	\textbf{8.20} \\
            \midrule
            MotionRNN~\cite{wu2021motionrnn} & 7.56M & 38.34G & $-$ & 0.921 &	31.99 &	137.77 &	8.79  \\
            \textbf{MS-MotionRNN} & 7.56M & \textbf{17.87G} & \textbf{-53.4\%} & \textbf{0.929} & \textbf{32.82} & \textbf{134.03} & \textbf{7.09} \\
            \bottomrule
		\end{tabular}
	} 
	\label{table:ms-rnn:kth-all}
\end{table}

\begin{figure}[htbp]
	\centering
	\includegraphics[width=0.6\linewidth]{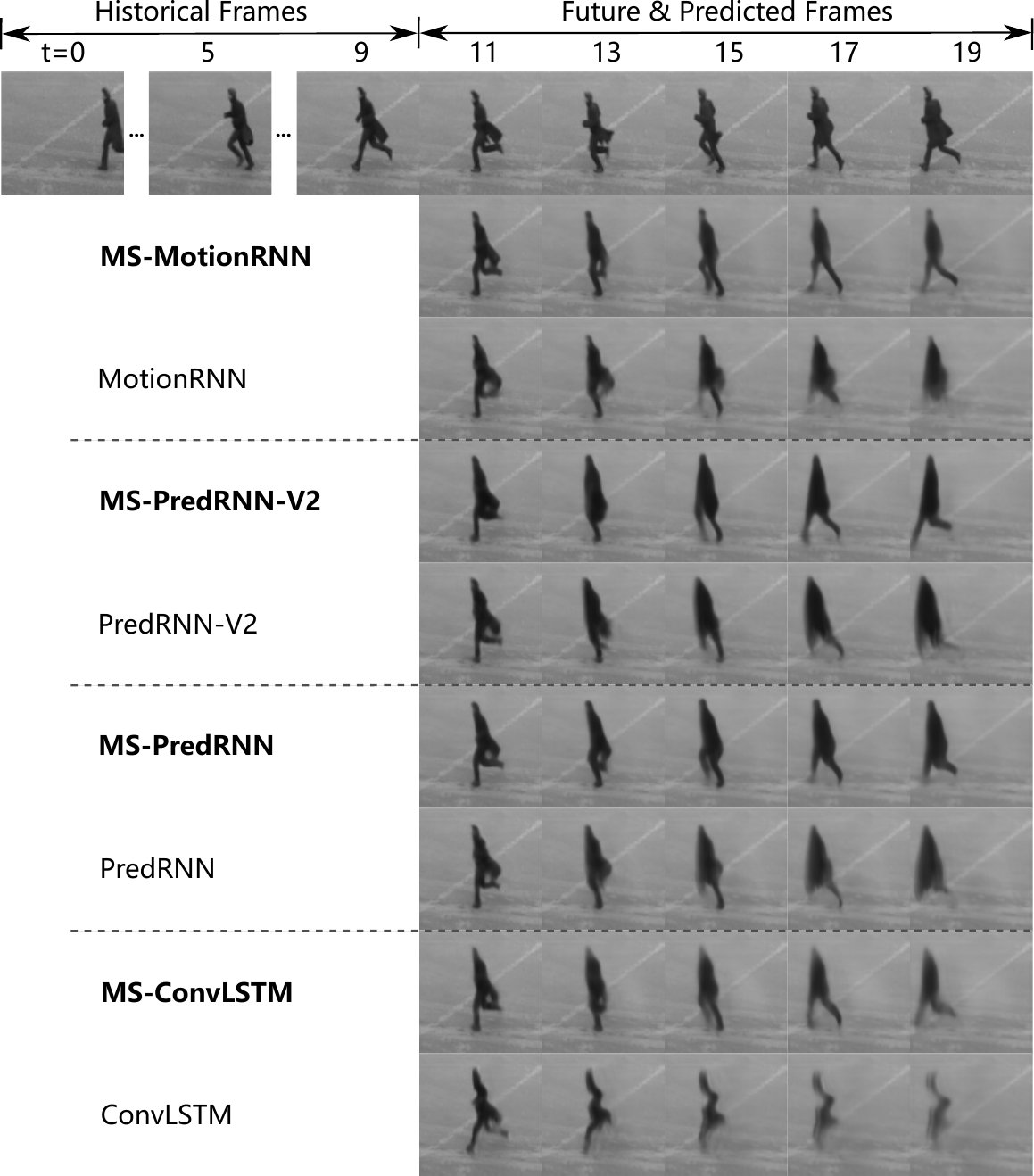}
	\caption{Qualitative comparison on the KTH dataset. RNN generates blurry and distorted predictions, while MS-RNN generates clear and natural predictions.}
	\label{fig:ms-rnn:kth_demo}
	\vspace{-0.5cm}
\end{figure}

\subsubsection{Visualization of Layer Outputs}  \label{sec:ms-rnn:layer out}
Similar to Fig.~\ref{fig:ms-rnn:minist_layer_out}, we also print the layer outputs of MS-ConvLSTM and ConvLSTM on the KTH dataset. By observing Fig.~\ref{fig:ms-rnn:kth_layer_out}, we get the same conclusion as Fig.~\ref{fig:ms-rnn:minist_layer_out}. The encoder of MS-ConvLSTM continuously captures motion by rapidly expanding the receptive field, and the decoder of MS-ConvLSTM gradually recovers the resolution from coarse to fine. This benefits from the design of multi-scale features. On the contrary, the encoder of ConvLSTM slowly expands the receptive field, and its decoder does not show a coarse-to-fine generation process. We can even clearly tell the human in all the feature maps of ConvLSTM, which shows that ConvLSTM has not learned advanced features, contrary to the fact that there are abstract features in the advanced layers of common CNNs.

\subsubsection{Comparison of the Training Cost and Performance}  \label{kth dingliang dingxing}
We adopt MSE, gradient difference loss (GDL)~\cite{mathieu2016deep}, peak signal to noise ratio (PSNR), and SSIM to perform the quantitative evaluation. Table~\ref{table:ms-rnn:kth-all} once again proves the previous inference that framework MS-RNN has the ability to reduce the training memory occupation of RNN and improve its performance. Fig.~\ref{fig:ms-rnn:kth_demo} exhibits a difficult prediction task, a man in a black windbreaker runs rapidly throughout the time window. In general, models with multiple scales yield better predictions than models without them. In detail, ConvLSTM, PredRNN, PredRNN-V2, and MotionRNN produce ambiguous and distorted predictions, while MS-ConvLSTM, MS-PredRNN, MS-PredRNN-V2, and MS-MotionRNN produce clear and natural predictions. However, the performance of MS-RNN is not completely satisfactory. For example, MS-MotionRNN gradually loses the prediction of the windbreaker; MS-ConvLSTM and MS-PredRNN-V2 wrongly predict the position of the upper body of the man in the later stage, it should be leaning forward instead of leaning back. Furthermore, none of the four of them captures the motion of the upper limbs. However, the MS-RNN's predictions are almost correct for both legs, which is where the most intense movement occurs. This also indirectly shows that MS-RNN has the ability to model fast motion.

\iffalse
\begin{figure}[htbp]
	\centering
	\subfigure[CSI-30]{\includegraphics[width=0.49\linewidth]{img/csi.pdf}}
    \subfigure[HSS-30]{\includegraphics[width=0.49\linewidth]{img/hss.pdf}}
	\caption{Framewise CSI-30 and HSS-30 comparison on the Germany dataset.}
	\label{fig:ms-rnn::csi-hss}
\end{figure}
\fi

\begin{table*}[htbp]
    \centering
	\caption{Training cost comparison and quantitative comparison on the Germany dataset.}
	\resizebox{1\linewidth}{!}{
		\centering
		\begin{tabular}{lccccccccccccccc}
			\toprule
			%\hline
			\centering
            Models & Params  & Memory & $\bigtriangleup$ & CSI-0.5$\uparrow$ & CSI-2$\uparrow$ & CSI-5$\uparrow$ & CSI-10$\uparrow$ & HSS-0.5$\uparrow$ & HSS-2$\uparrow$ & HSS-5$\uparrow$ & HSS-10$\uparrow$ & B-MSE$\downarrow$ & B-MAE$\downarrow$ \\
            \midrule
            %\hline
            ConvLSTM~\cite{shi2015convolutional} & 1.77M & 7.86G & $-$ & 0.390 & 0.219 & 0.095 & 0.002 & 0.539 & 0.333 & 0.162 & 0.003 & 92.54 & 290.85 \\
            \textbf{MS-ConvLSTM} & 1.77M &	\textbf{5.02G} & \textbf{-36.1\%} & \textbf{0.407} & \textbf{0.235} & \textbf{0.103} & \textbf{0.008} & \textbf{0.560} & \textbf{0.359}  & \textbf{0.176}  & \textbf{0.015} & \textbf{86.54} & \textbf{280.96} \\
            \midrule
            PredRNN~\cite{wang2017predrnn} & 3.59M & 17.01G & $-$ & 0.414 &	0.238 &	0.096 &	0.006 &	0.565 &	0.361 & 0.164 & 0.011 & 84.72 & 278.33 \\
            \textbf{MS-PredRNN} & 3.59M & \textbf{9.17G} & \textbf{-46.1\%} & \textbf{0.427} & \textbf{0.256} &	\textbf{0.107} & 0.006 & \textbf{0.580} & \textbf{0.389} & \textbf{0.183} & \textbf{0.012} & \textbf{81.63}  & \textbf{272.63}\\
            \midrule
            MIM~\cite{wang2019memory} & 7.36M & 32.75G  & $-$ & 0.418 & 0.253 & 0.124 &	0.007 &	0.568 &	0.379 & 0.207 & 0.014 & 84.23 & 275.24 \\
            \textbf{MS-MIM} & 7.36M & \textbf{16.06G} & \textbf{-51.0\%} & \textbf{0.427} &	\textbf{0.277} & \textbf{0.134} &	\textbf{0.013} & \textbf{0.579} & \textbf{0.415} & \textbf{0.223} & \textbf{0.026} & \textbf{81.77} & \textbf{273.82} \\ 
            \midrule
            MotionRNN~\cite{wu2021motionrnn} & 7.56M & 35.28G & $-$ & 0.431 &	0.259 &	0.103 &	0.001 &	0.583 &	0.388 & 0.176 & 0.002 & 81.24 & 270.73 \\
            \textbf{MS-MotionRNN} & 7.56M &	\textbf{17.27G} & \textbf{-51.1\%} & \textbf{0.437} & 	\textbf{0.268} & \textbf{0.116} & \textbf{0.007} & \textbf{0.592} & \textbf{0.406} & \textbf{0.198} & \textbf{0.013} & \textbf{80.62} & \textbf{269.35} \\
            \midrule
            PrecipLSTM~\cite{ma2022preciplstm} & 7.04M & 40.00G & $-$ & 0.426 &	0.264 &	0.110 &	0.003 &	0.577 &	0.395 & 0.185 & 0.006 & 82.10 & 272.73 \\
            \textbf{MS-PrecipLSTM} & 7.04M & \textbf{19.48G} & \textbf{-51.3\%} & \textbf{0.435} & 	\textbf{0.272} & \textbf{0.118} & \textbf{0.007} & \textbf{0.588} & \textbf{0.409} & \textbf{0.202} & \textbf{0.015} & \textbf{80.43} & \textbf{268.36} \\
            \bottomrule
		\end{tabular}
	} 
	\label{table:ms-rnn:hko-cost}
	\vspace{-0.2cm}
\end{table*}

\subsection{Precipitation Nowcasting}\label{sec:ms-rnn:hko}
The temporal span of the Germany radar dataset~\cite{ayzel2020rainnet} is from 2006 to 2017, and the spatial span is the entire German continent. The dataset is collected every 5 minutes by 17 Doppler radars. Data from 2006 to 2014 are used for training and data from 2015 to 2017 are used for testing. We interpolate radar maps to 124$\times$124 using bilinear interpolation and sample a frame every 30 minutes. The model needs to predict 5 unknown frames based on 5 known frames, i.e., the precipitation forecast duration is 2.5 hours.

\begin{figure}[htbp]
	\centering                                     % 78  90  165
	\includegraphics[width=0.6\linewidth]{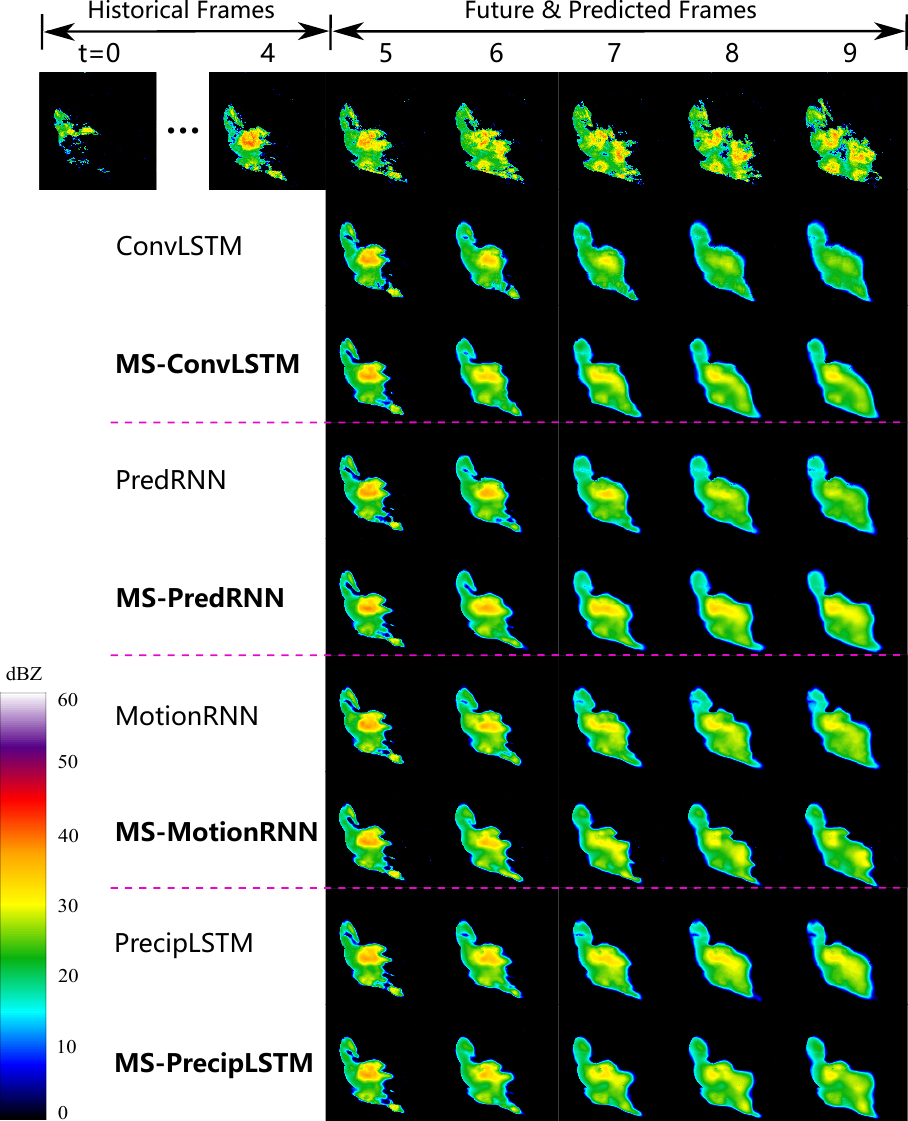}
	\caption{Qualitative comparison on the Germany dataset. The prediction of MS-RNN pays more attention to the precipitation higher than 3 $mm/h$ (yellow, 30dBZ), which is more harmful.}
	\label{fig:ms-rnn:hko_demo}
	\vspace{-0.5cm}
\end{figure}

\subsubsection{Comparison of the Training Cost and Performance}  \label{germany dingliang dingxing}

From Table~\ref{table:ms-rnn:hko-cost} we can see that the memory occupation of the state-of-the-art model PrecipLSTM has reached 40G, the maximum capacity of the A100 graphics card. However, the size of the image processed by PrecipLSTM at this time is only 124$\times$124, which is far from the high-resolution requirements of precipitation nowcasting in practical applications. However, MS-RNN greatly alleviates this dilemma. MS-MIM, MS-MotionRNN, and MS-PrecipLSTM reduce the memory usage of the basic RNN model by nearly half.

Both critical success index (CSI)~\cite{shi2017deep} and Heidke skill score (HSS)~\cite{shi2017deep} need to be calculated under a certain threshold, and this paper takes 0.5, 2, 5, and 10 $mm/h$ as the threshold of rainfall intensity. We ignore indicators for heavy rain because Germany has a temperate maritime climate with frequent light rain all year round. Furthermore, we also evaluate the overall performance of the model with the balanced mean squared error (B-MSE)~\cite{shi2017deep} and balanced mean absolute error (B-MAE)~\cite{shi2017deep} measures, which assign more weights to heavier rainfalls in the calculation of MSE and mean absolute error (MAE).

Table~\ref{table:ms-rnn:hko-cost} gives quantitative results. MS-RNN improves the precipitation forecasting ability of RNN, among which the performance of MS-PredRNN is close to the advanced PrecipLSTM, while MS-ConvLSTM has average performance but the smallest memory usage. In practical applications, MS-ConvLSTM and MS-PredRNN can be used to cope with high-resolution prediction tasks. This paper adopts an RNN structure of 64 channels and 6 layers, which also can be appropriately reduced to find a trade-off between performance and cost.

Fig.~\ref{fig:ms-rnn:hko_demo} also proves that MS-RNN has better performance than RNN. MS-RNN pay more attention to the precipitation over 30 $\textrm{dBZ}$ (yellow, which can be converted to about 3 $mm/h$ through the Z-R relationship in Germany~\cite{ma2022preciplstm}) than RNN, which is most evident in their predictions for the last frame.

\subsection{Ablation and Further Experiments}

\subsubsection{Ablation Studies}  \label{ablation}
\begin{table}[htbp]
    \centering
	\caption{Ablation study on the Moving MNIST dataset. The symbols $w$ and $w/o$ represent with and without respectively. MS and Skip are short for multi-scale and skip connections.}
	\resizebox{0.6\linewidth}{!}{
		\centering
		\begin{tabular}{lccccc}
			\toprule
			%\hline
			\centering
            Models & Parts & Params & Memory & FLOPs & MSE$\downarrow$ \\
            \midrule
            ConvLSTM & $w/o$ MS, $w/o$ Skip & 1.77M & 5.50G & 137.7G & 82.0   \\
            MS-ConvLSTM & $w$ MS, $w/o$ Skip & 1.77M & 3.86G & 60.3G & 62.8   \\
            MS-ConvLSTM  & $w$ MS, $w$ Skip & 1.77M & 3.86G & 60.3G & 58.3    \\
            \bottomrule
		\end{tabular}
	} 
	\label{table:ms-rnn:mnist-ablation}
    \vspace{-0.4cm}
\end{table}

%We use ConvLSTM as the basic model to experiment with the effect of multi-scale structure and skip connection (UNet way). From Table~\ref{table:ms-rnn:mnist-ablation}, we can see that ConvLSTM's performance has been improved with the addition of them in sequence, and the memory and FLOPs usage has been reduced with the introduction of the multi-scale structure. In addition, it can be seen that skip connections (``+'') do not introduce additional parameters, memory, and FLOPs (almost negligible) from the comparison of the last two rows.

The purpose of ablation studies is to verify whether the two important components (the multi-scale structure and skip connections (UNet way)) introduced by MS-RNN on the basis of RNN are effective. We use ConvLSTM as the base model and experiment on the Moving MNIST dataset. The results are shown in Table~\ref{table:ms-rnn:mnist-ablation}. From the comparison of the first two rows in the table, it can be seen that the introduction of multi-scale design has brought about the reduction of memory, FLOPs, and MSE, which shows that the introduction of multi-scale design is beneficial and bring low consumption and high performance. In addition, MS-ConvLSTM maintains the same parameters as ConvLSTM, which means that sampling layers do not bring additional parameters. From the comparison of the last two rows in the table, it can be seen that the introduction of skip connections (``+'') has brought about a further reduction in MSE, which shows that the introduction of skip connection is also beneficial, which can preserve high-frequency information and avoid information loss caused by sampling. In addition, MS-ConvLSTM with skip connections maintains the same parameters, memory, and FLOPs as MS-ConvLSTM without skip connections, which means that skip connections do not bring additional parameters, memory, and FLOPs. In theory, the ``+'' operation will bring extra FLOPs, but the few operations used by MS-ConvLSTM are almost negligible.

\begin{table}[htbp]
    \centering
	\caption{Comparison of different skip connection ways on the Moving MNIST dataset.}
	\resizebox{0.55\linewidth}{!}{
		\centering
		\begin{tabular}{lccccc}
			\toprule
			%\hline
			\centering
            Models & Ways & Params & Memory & FLOPs & MAE$\downarrow$ \\
            \midrule
            MS-ConvLSTM & FC & 1.77M & 3.88G & 60.5G & 110.4 \\
            MS-ConvLSTM & UNet 3+ & 1.77M &	3.87G & 60.4G &	109.5    \\
            \textbf{MS-ConvLSTM}  & \textbf{UNet} & 1.77M & \textbf{3.86G} & \textbf{60.3G} & \textbf{109.2}   \\
            \bottomrule
		\end{tabular}
	} 
	\label{table:ms-rnn:mnist-skip}
    \vspace{-0.4cm}
\end{table}

\subsubsection{Further Experiments on Skip Connection Ways}  \label{skip ways}
Furthermore, we do more in-depth experiments on the skip connection settings using MS-ConvLSTM on the Moving MNIST dataset, such as similar to UNet~\cite{ronneberger2015u}, UNet 3+~\cite{huang2020unet}, and fully connected (FC) ways. The difference between the three models is that they have different numbers of sampling and skip connection (``+'') operations, among which MS-ConvLSTM (FC) is the most, MS-ConvLSTM (UNet) is the least, and MS-ConvLSTM (UNet 3+) is in the middle. Table~\ref{table:ms-rnn:mnist-skip} exhibits the experimental results. First, it goes without saying that neither sampling nor skipping connections have parameters, so the three have the same parameters. Second, although skip connections do not have any memory footprint~\cite{sohoni2019low}, sampling has output memory~\cite{gao2020estimating}, so MS-ConvLSTM (UNet) has the least memory footprint. Third, both sampling and skip connections have FLOPs, so MS-ConvLSTM (UNet) has the least FLOPs. Finally, MS-ConvLSTM (UNet) has the least MAE. Overall, the same-scale skip connection between the encoder and decoder like UNet is the simplest but the most effective.

\begin{table}[htbp]
    \centering
	\caption{Comparison of different losses on the Moving MNIST dataset.}
	\resizebox{0.55\linewidth}{!}{
		\centering
		\begin{tabular}{lccccc}
			\toprule
			%\hline
			\centering
            Models & Losses & SSIM$\uparrow$ & GDL$\downarrow$ & MSE$\downarrow$ & MAE$\downarrow$\\
            \midrule
            MS-ConvLSTM & $L_1$ & 0.852 & 171.8 & 71.3 & 
111.4 \\
            MS-ConvLSTM & $L_2$ & 0.844 & 172.7 & \textbf{49.5} & 122.3 \\
            \textbf{MS-ConvLSTM}  & \bm{$L_1 + L_2$} & \textbf{0.863} & \textbf{167.4} & 58.3 & \textbf{109.2} \\
            \bottomrule
		\end{tabular}
	} 
	\label{table:ms-rnn:loss vs}
    \vspace{-0.4cm}
\end{table}

\subsubsection{Further Experiments on Loss Functions}  \label{loss func}

Previous works use different loss functions, such as using only $L_1$~\cite{byeon2018contextvp} or $L_2$~\cite{wang2019memory, fan2019cubic, wang2022predrnn} or a combination of the two~\cite{wang2017predrnn, wang2018predrnn++}, but they do not explain the reason. The MSE loss (or $L_2$) is derivable everywhere, and the gradient is dynamically changed, enabling the model to converge quickly. However, the square term will amplify the loss of outliers~\cite{mathieu2016deep}, at the cost of sacrificing the errors of other samples, and the parameters will be updated in the direction of reducing the errors of outliers, which will reduce the overall performance of the model. In contrast, MAE loss (or $L_1$) is softer in dealing with outliers~\cite{mathieu2016deep}, but it has other problems: the gradient is always large during training, and it is continuous but non-differentiable at 0, which means that its gradient is large even for a small loss. This is obviously not conducive to the learning of the model. A common practice is to use a combination of MSE and MAE to optimize the model, which is also adopted by MS-RNN. Table~\ref{table:ms-rnn:loss vs} also proves this point, the combination of $L_1$ and $L_2$ is better than using only one of them. However, the pixel-wise loss like $L_p$ copes with future uncertainties simply by averaging pixels, leading to blurry predictions, which run counter to human visual perception. Some works introduce regularization terms on the basis of $L_p$ loss to alleviate the blurring problem, such as SSIM~\cite{chang2021mau, lotter2017deep}, GDL~\cite{mathieu2016deep, villegas2017decomposing}, perceptual loss~\cite{chang2022strpm}, adversarial loss~\cite{ravuri2021skilful, chang2022strpm}, and KL divergence loss~\cite{denton2018stochastic, akan2021slamp, wu2021greedy}, etc. Regrettably, the problem of ambiguity has not been completely solved. The regress-to-the-mean problem remains an open issue~\cite{oprea2020review}.

\section{Discussion} \label{discussion}
Based on the extensive experimental results in this paper, we summarize and discuss some knowledge derived in the context of spatiotemporal prediction, which may be beneficial to future research.

\subsection{How to choose or design video prediction models?}

Video prediction is a pixel-level spatiotemporal prediction task, which is a tough issue. Whoever has grasped the essence of spatiotemporal motion can be competent for this task.

From the perspective of the choice of model architecture, the convolutional RNN model is more suitable for this task. The seamless integration of CNN and RNN perfectly adapts to the spatiotemporal characteristics of video, which is why their family members are numerous and enduring. UNet~\cite{ronneberger2015u} and SimVP~\cite{gao2022simvp} of the CNN family implicitly assume that convolutions can model spatiotemporal features. However, convolution is not suitable for processing time-series data, it is designed for extracting spatial features. This causes SimVP to spend a large number of epochs to force the convolution to learn the spatiotemporal characteristics of the data, which is not only time-consuming but also ineffective. Although the Earthformer~\cite{gao2022earthformer} of the Transformer family has considered the spatiotemporal characteristics of data and introduced local spatiotemporal attention when designing the network, which reduces the training difficulty of Transformer to a certain extent, the computational cost of attention is too high and it is not as effective as CNN and RNN. Moreover, complex 3D attention designs are also prohibitive.

From the perspective of model design, a model with a larger spatial receptive field and more temporal memories is more suitable for this task. If the convolutional RNN architecture has been chosen, then models with large spatial receptive fields and multiple temporal memories are better designs, like MS-MotionRNN or MS-PrecipLSTM. Spatiotemporal motion changes in spatial position, and if the model has a large spatial receptive field, then it can see and learn from these changes. Spatiotemporal motion changes over time, and if the model has more than one memory, then it can remember and learn from these changes.

\subsection{How to choose or design video prediction datasets?}

Video prediction is a broad problem with different applications in different domains. Therefore, the selection or production of datasets should cover as many of these subfields as possible, at least covering subfields with important practical application value, such as traffic flow prediction, pedestrian path prediction, weather forecast, etc. Of course, it is also feasible to use the synthetic dataset Moving MNIST for quick experiments. Another consideration that needs to be taken into account when selecting or designing a dataset is that there should be different rates of motion in the dataset. Natural movement is variable rather than fixed. Motions at different rates represent changes at different scales, and the model needs to be equipped with receptive fields of different scales to capture them, which increases the difficulty of prediction. From the generation of the Moving MNIST and KTH datasets, it can be intuitively seen that there are different rates of motion. The TaxiBJ and German datasets implicitly contain different rates of motion, since the movement speeds of vehicles and precipitation should also be random. The four datasets selected in this paper are comprehensive and unbiased, and they are recommended to test the performance of the model.

\section{Conclusion and Future Work}\label{conclusion and future work}
Previous RNNs obtain higher predictive ability by widening or deepening networks, but the side effect is a sharp increase in computing resources, especially video memory. Unlike them, this paper uses multi-scale technology to improve the model's performance, which also brings a significant reduction in memory. We unify these RNNs' structures and embed multi-scale architectures into them to propose a general framework called MS-RNN. We mathematically analyze the reduced ratio of memory and FLOPs brought by MS-RNN and try to explain the performance improvement brought by MS-RNN from the perspective of the receptive field and so on. We perform comparative experiments on four video datasets (Moving MNIST, TaxiBJ, KTH, and Germany) using eight base RNNs (ConvLSTM, TrajGRU, PredRNN, PredRNN++, MIM, MotionRNN, PredRNN-V2, and PrecipLSTM). The results show that MS-RNN is low-cost and high-performance. Specifically, experiments verify previous theoretical analysis: the reduction of memory and FLOPs brought by MS-RNN will not exceed 56.25\% and equal to 56.25\% respectively; MS-RNN does have a larger receptive field and predict from coarse to fine, which is the reason for the higher performance of MS-RNN. In addition, we also compare with the competing multi-scale RNN (CMS-LSTM and MoDeRNN), multi-scale CNN (SimVP), and multi-scale Transformer (Earthformer), and results show that MS-RNN is more competitive in memory and performance.

Although MS-RNN greatly reduces the memory usage of RNN, it does not completely solve the problem of training memory shortage. A way to further reduce memory usage is to use patch. Patch is to cut the input image into pieces, stitch these pieces together in the channel dimension for input into the model, and finally restore the model output with the reverse cutting operation. ConvLSTM first adopts patch to alleviate the tight memory usage. However, using patch can lead to a significant drop in model performance and may cause the tilling effect. In addition, the introduction of patch is the fundamental reason why the Transformer architecture can be transferred from natural language processing to vision (ViT), which reduces the space complexity of Transformer. In conclusion, patch can reduce the training memory footprint, but the balance between prediction cost and prediction skill needs further study.

\section{Acknowledgments}
This work is partly supported by the National Key R\&D Program of China under Grant No. 2021ZD0110900, the National Natural Science Foundation of China under Grant No. 62106061, 61972114, the Fundamental Research Funds for the Central Universities under Grant No. FRFCU5710010521, the Research and Application of Key Technologies for Intelligent Farming Decision Platform, An Open Competition Project of Heilongjiang Province (China) under Grant No. 2021ZXJ05A03, the Key Research and Development Program of Heilongjiang Province under Grant No. 2022ZX01A22, the National Natural Science Foundation of Heilongjiang Province under Grant No. YQ2019F007.

\bibliography{ms-rnn}
\bibliographystyle{elsarticle-num}   % 数字引用-有序的

\end{sloppypar}
\end{document}